\documentclass[sigconf]{acmart}

\AtBeginDocument{%
  \providecommand\BibTeX{{%
    \normalfont B\kern-0.5em{\scshape i\kern-0.25em b}\kern-0.8em\TeX}}}

\copyrightyear{2022}
\acmYear{2022}
\setcopyright{rightsretained}
\acmConference[ASIA CCS '22]{Proceedings of the 2022 ACM Asia Conference on Computer and Communications Security}{May 30-June 3, 2022}{Nagasaki, Japan}
\acmBooktitle{Proceedings of the 2022 ACM Asia Conference on Computer and Communications Security (ASIA CCS '22), May 30-June 3, 2022, Nagasaki, Japan}\acmDOI{10.1145/3488932.3517406}
\acmISBN{978-1-4503-9140-5/22/05}

\usepackage{booktabs}
\usepackage{amsfonts}
\usepackage{bm}
\usepackage{makecell}
\usepackage{algorithm}
\usepackage{algorithmicx}
\usepackage{algpseudocode}
\usepackage{url}
\usepackage{enumitem}

\usepackage{xcolor}

\newcommand{\shortname}{{MCMC-based machine unlearning}}
\newcommand{\acron}{{MCU}}
\newcommand{\error}{{unwanted}}

\theoremstyle{definition}

\newtheorem{remark}{Remark}

\newcommand{\argmax}{\arg\!\max}
\newcommand{\argmin}{\arg\!\min}
\newcommand{\mbf}[1]{\mathbf{#1}}
\newcommand{\mcl}[1]{\mathcal{#1}}
\newcommand{\mbb}[1]{\mathbb{#1}}

\begin{document}

\title{Markov Chain Monte Carlo-Based Machine Unlearning: Unlearning What Needs to be Forgotten}

\author{Quoc Phong Nguyen}
\affiliation{
	\institution{National University of Singapore}
	\country{}
}
\email{qphong@comp.nus.edu.sg}

\author{Ryutaro Oikawa}
\affiliation{
	\institution{National University of Singapore}
	\country{}
}
\email{ryutaro.oikawa.1991@gmail.com}

\author{Dinil Mon Divakaran}
\affiliation{
	\institution{Trustwave}
	\country{}
}
\email{dinil.divakaran@trustwave.com}

\author{Mun Choon Chan }
\affiliation{
	\institution{National University of Singapore}
	\country{}
}
\email{chanmc@comp.nus.edu.sg}

\author{Bryan Kian Hsiang Low }
\affiliation{
	\institution{National University of Singapore}
	\country{}
}
\email{lowkh@comp.nus.edu.sg}

\renewcommand{\shortauthors}{Q. P. Nguyen, R. Oikawa, D. M. Divakaran, M. C. Chan, and B. K. H. Low}

\begin{abstract}
As the use of machine learning (ML) models is becoming increasingly popular in many real-world applications, there are practical challenges that need to be addressed for model maintenance. One such challenge is to `undo' the effect of a specific subset of dataset used for training a model. This specific subset may contain malicious or adversarial data injected by an attacker, which affects the model performance. Another reason may be the need for a service provider to remove data pertaining to a specific user to respect the user's privacy. In both cases, the problem is to `unlearn' a specific subset of the training data from a trained model without incurring the costly procedure of retraining the whole model from scratch. Towards this goal, this paper presents a Markov chain Monte Carlo-based machine unlearning (\acron) algorithm.
MCU helps to effectively and efficiently unlearn a trained model from subsets of training dataset. 
Furthermore, we show that with \acron{}, we are able to explain the effect of a subset of a training dataset on the model prediction. Thus, MCU is useful for examining subsets of data to identify the adversarial data to be removed.
Similarly, \acron{} can be used to erase the lineage of a user's personal data from trained ML models, thus upholding a user’s ``right to be forgotten''.
We empirically evaluate the performance of our proposed  \acron{} algorithm on real-world phishing and diabetes datasets. Results show that \acron{} can achieve a desirable performance by efficiently removing the effect of a subset of training dataset and outperform an existing algorithm that utilizes the remaining dataset.
\end{abstract}



\maketitle

\section{Introduction}
\label{sec:introduction}

While there has been a rapid increase in machine learning (ML) applications, they often require accurately labeled datasets to achieve competitive performance. This is also common in the security domain where supervised classifiers are built for threat detection (e.g., for detecting phishing attacks~\cite{le2018urlnet, lee2020building, marchal2016know, visualphishnet-2020, phishpedia-2021, d-fence-2021}, 
malware and malicious communications~\cite{graph-mw-dyn-2011, mw-detection-MALWARE-2015, enc-mw-traffic-2016, NADA-2018, EMBER-2018, nguyen2019gee,  MaMaDroid-2019}). However, even with significant and costly human verification~\cite{phishtank-eval-2008, fireEye-ML-mw-2018}, these datasets are prone to errors and poisoning attacks. 
For example, in the case of phishing, attackers often host the phishing pages on compromised sites~\cite{phishing-compromised-2021} which are taken down and handed over to the website owners when detected.
Thus, due to the short life cycle of phishing attacks~\cite{phish-BL-analysis-2020, phishing-life-cycle-2020}, not all URLs gathered from a feed (e.g., OpenPhish~\cite{openphish} or PhishTank~\cite{phishtank}) correspond to valid phishing webpages. Such errors inadvertently corrupt the dataset with wrong labels. This undesirable 
outcome
can also be caused by an adversary that deliberately injects malicious data into a training dataset~\cite{szegedy2013intriguing,goodfellow2014explaining,koh2017understanding} (e.g., see~\cite{backdoor-poisoning-malware-2021} for backdoor poisoning attack against malware classifiers), thereby degenerating the model performance.
A straightforward solution is to retrain the model from scratch after removing the \error\footnote{We also refer to the unwanted dataset as the \emph{erased dataset}.} (erroneous or malicious) data from the training dataset. But, such an approach is not practical as it incurs a  significant amount of time and storage, especially when the \emph{remaining dataset} (after removing the \error{} data) is large. Besides, as discussed below, since user privacy is of paramount importance, retaining data is not an option in all use cases~\cite{GDPR-right-to-be-forgotten}.

While the above settings assume that the set of \error{} data is given, there is also a related and important problem of identifying the subset of training data that is malicious (or erroneous) by examining a number of subsets of data. This problem arises when the training dataset is formed using different subsets acquired from multiple sources: For example, phishing URLs and malicious domains can be obtained from multiple sources such as OpenPhish~\cite{openphish}, PhishTank~\cite{phishtank}, APWG~\cite{APWG}, URLhaus~\cite{urlhaus}, Anomali~\cite{anomali}, and other commercial as well as open source threat intelligence feeds. Among these subsets of training data, the objective is to examine whether there exists a subset of malicious data. 
We consider the approach of analyzing the model prediction after removing a subset of the training data. In particular, a subset of the training data is potentially malicious if removing it from the model increases the accuracy in the model prediction (e.g., on a test set). Due to the vast number of the subsets of a training dataset to be examined, it is not practical to retrain the model numerous times to identify the effect of removing each subset on the model prediction.
This problem is also important when we want to apportion credit between different network analysts that label different parts of the data.

The problem of removing a subset of the training dataset from a trained model also arises in the domain of user privacy. In particular, a company or service provider needs to remove a user's personal data when she would like to exercise the ``right to be forgotten''~\cite{right-to-be-forgotten-AI-2018}. The data points are not only stored in the database (which can be easily deleted), but they are also used to train ML models, thus forming the \emph{data lineage}~\cite{cao2015towards}. Therefore, to respect the privacy of a user, it is important that her data's lineage is also erased from the trained ML models upon request. Since the number of users who demand their data to be removed (e.g., those who stop using a service from a company) is often much smaller than that of the remaining users (e.g., active users of the service), retraining the model from scratch by excluding the erased data is prohibitively expensive. Furthermore, in some cases, the majority of the user data may not be accessible due to storage limitations or policies restricting the retention period of data.

Note that, while there exist approaches to learn with noisy labels~\cite{angluin1988learning,patrini2017making,goldberger2017}, they address a problem different from what we have discussed above.
The above scenarios highlight a common problem of removing the effect of a subset of the training dataset from a model without retraining the model from scratch, which is called \emph{machine unlearning} (MU) \cite{cao2015towards,du2019lifelong,ginart2019making,garg2020formalizing,nguyen2020variational}. When the accuracy of the model after being unlearned is not crucial, the approach of analyzing the model prediction after removing a subset of the training data is also investigated in the \emph{interpretable ML} literature~\cite{koh2017understanding,koh2019accuracy}.

\textbf{Contribution:} In this work, we address the problem of unlearning, 
under the assumption that there are samples of the model parameters which estimate the posterior belief of the model parameters well. In particular, we employ a \emph{Markov chain Monte Carlo} (MCMC) algorithm to draw these samples. 
Hence, our novel proposed approach is named \emph{\shortname{}} (\acron). It can efficiently unlearn a trained model from different subsets of the training dataset.
The key component of \acron{} is a \emph{candidate set} of the model parameters such that the parameters of the model retrained on the remaining dataset (i.e., the naively \emph{retrained model}) are close to a candidate in the candidate set (Section~\ref{sec:candidateset}). 

We also assume that the subset of the training data to be erased is small relative to the whole training dataset, which means that the change in model parameters by removing the subset is small. Hence, the posterior probability of the retrained model parameters given the training dataset should be sufficiently large. This is the rationale behind our selection of the candidate set as the set of MCMC samples drawn from the posterior belief of the model parameters given the training dataset. The candidate set is equipped with auxiliary values based on the importance sampling technique that are useful in performing unlearning (Section~\ref{sec:unlearnwithcandidate}).
We further relax our assumption on the small change of the model parameters by proposing an \emph{enlarged candidate set} based on a flattened distribution (Section~\ref{sec:enlargecandidateset}). Additionally, \acron{} can be used to explain the effect of a subset of training data on the model prediction (Section~\ref{sec:datainfluence}). 
We also show that \acron{} does not suffer from an important pitfall of \emph{catastrophic unlearning} (Section~\ref{sec:pitfall}) known to affect MU algorithms.

We empirically demonstrate the advantage of \acron{} with the enlarged candidate set in three scenarios---removing user's personal data, removing \error{} data, and interpreting the model prediction with respect to subsets of the training data.
Note, since MCMC faces difficulty in high dimensional problems~\cite{barbos17,betancourt2017conceptual}, our experiments are conducted on problems of moderate dimensions where the above assumption holds, i.e., the MCMC samples estimate the posterior distribution of the model parameters well.
Nonetheless, we perform extensive experiments using both synthetic and real-world datasets, the latter of which include 
the Pima Indians diabetes dataset and a phishing webpage detection dataset (Section~\ref{sec:experiments}). Results show that our proposed enlarged candidate set improves the performance of \acron{} significantly. Furthermore, while not using the remaining dataset during unlearning, \acron{} with an enlarged candidate set can outperform an existing MU algorithm \cite{fu2021bayesian,fu22} that however utilizes the remaining dataset in its unlearning procedure. To the best of our knowledge, \acron{} is the only work on MU for MCMC algorithms without using the remaining dataset.

\section{Related Works}

The pioneering work~\cite{cao2015towards} addresses the MU problem for statistical query learning \cite{kearns1998efficient} which includes several ML algorithms such as naive Bayes classifier, support vector machine, and $k$-means clustering. However, they are required to have a summation form.
To perform unlearning, the erased data are simply subtracted from the summations and the model is updated.
The work of~\cite{bourtoule2019machine} proposes to divide the training dataset into multiple shards, each of which is trained with a different model. By assuming that the erased dataset belongs to a small number of shards, the computation required in retraining models is reduced.
Another class of research works~\cite{guo2019certified,fu2021bayesian,fu22} utilize the notion of an influence function \cite{cook1982residuals,koh2017understanding} to unlearn a model from a datum (i.e., an individual data point). In particular, the work of~\cite{guo2019certified} can unlearn logistic regression models, while that of~\cite{fu2021bayesian,fu22} can unlearn Bayesian models such as those obtained from MCMC algorithms. However, since the influence function is based on the first-order Taylor approximation, these unlearning algorithms are only accurate when there is a small change in the model (e.g., erasing a datum from a large training dataset), which makes them less practical for real-world use.
Other works focus on unlearning specific learning algorithms such as $k$-means clustering \cite{ginart2019making} and variational inference \cite{nguyen2020variational}.

\section{Background}
\label{sec:background}

\subsection{Machine Unlearning}
\label{sec:machineunlearning}

We focus on supervised learning problems where the training dataset consists of input and class-label pairs. ML models are used to capture the conditional probability $p(y|\mbf{x}, \bm{\theta})$ of a label $y$ given an input $\mathbf{x}$, and the model parameters $\bm{\theta}$. Let the training data be denoted by $\mcl{D}~\triangleq~\{(\mbf{x}_i,y_i)\}_{i=1}^n$. Given the training dataset, we would like to find the \emph{maximum a posteriori} (MAP) estimate of the model parameters. In other words, the optimal model parameters $\bm{\theta}_{\mcl{D}}$ maximize the \emph{posterior} probability $p(\bm{\theta}|\mcl{D})$ of the model parameters given the training data:
\begin{equation}
\bm{\theta}_{\mcl{D}} \hspace{-0.5mm}\triangleq\hspace{-0.5mm} \mathop{\argmax}_{\bm{\theta}} \log p(\bm{\theta}|\mcl{D})
    \hspace{-0.5mm}= \hspace{-0.5mm}\mathop{\argmax}_{\bm{\theta}} \left( \log p(\mcl{D}|\bm{\theta}) + \log p(\bm{\theta}) \right)
\label{eq:map}
\end{equation}
where $p(\mcl{D}|\bm{\theta})$ and $p(\bm{\theta})$ are the \emph{likelihood} and the \emph{prior}, respectively.
The 
data points are assumed to be conditionally independent given the model parameters, which holds for many ML models such as regression models and feed-forward neural networks: 
\begin{equation}
\log p(\mcl{D}|\bm{\theta}) = \log \prod_{i=1}^n  p(y_i|\mbf{x}_i, \bm{\theta}) = \sum_{i=1}^n \log p(y_i|\mbf{x}_i, \bm{\theta})\ .
\label{eq:condindep}
\end{equation}

Let us consider the \emph{machine unlearning} scenario where we would like to remove the effect of an \emph{erased dataset}, denoted by $\mcl{D}_e \subset \mcl{D}$, from a trained model specified by $\bm{\theta}_{\mcl{D}}$ \cite{cao2015towards}. Recall that $\bm{\theta}_{\mcl{D}}$ is the MAP estimate of the model parameters given the training data $\mcl{D}$. It is obtained by maximizing the log posterior probability of $\bm{\theta}$ given $\mcl{D}$ (Eq.~\eqref{eq:map}). When the erased dataset $\mcl{D}_e$ is relatively large compared to the \emph{remaining dataset}, denoted by $\mcl{D}_r \triangleq \mcl{D} \setminus \mcl{D}_e$ (obtained by removing the erased dataset $\mcl{D}_e$ from $\mcl{D}$), retraining the model from scratch using the remaining dataset $\mcl{D}_r$ is a possible solution, i.e., finding 
\begin{equation*}
\bm{\theta}_{\mcl{D}_r} \triangleq \mathop{\argmax}_{\bm{\theta}} \log p(\bm{\theta}| \mcl{D}_r)\ .
\end{equation*} 
In most practical scenarios, the erased dataset $\mcl{D}$ is small relative to the size of entire dataset $\mcl{D}$. 
For example, whether it is applying unsupervised deep learning models to detect anomalies in network traffic~\cite{nguyen2019gee} or 
removal of user data (lineage) from existing trained models, 
$\mcl{D}_r$ is typically much larger than $\mcl{D}_e$. 
Therefore, retraining the model on $\mcl{D}_r$ from scratch can be inefficient in terms of computational time and impractical in terms of storing data indefinitely due to both storage constraints and/or regualatory policies. Also, note that when $\mcl{D}_e$ is small, the difference between the parameters $\bm{\theta}_{\mcl{D}}$ of the model trained on $\mcl{D}$ vs.~the parameters $\bm{\theta}_{\mcl{D}_r}$ of the model trained on $\mcl{D}_r$ is less drastic. Hence, MU algorithms can achieve a more efficient solution without resorting to the naive approach of retraining with the remaining data $\mcl{D}_r$~\cite{cao2015towards}.

Given the likelihood function and the prior distribution in Eq.~\eqref{eq:map}, let us review a stochastic approximation method of the posterior distribution $p(\bm{\theta}|\mcl{D})$ of the model parameters given the training dataset called the \emph{Markov chain Monte Carlo} (MCMC) sampling. It will be used to construct the \emph{candidate set} in our proposed \acron{} algorithm later in Section~\ref{sec:candidateset}.

\subsection{Markov Chain Monte Carlo}
\label{sec:mcmc}

\emph{Markov chain Monte Carlo} (MCMC) algorithms are used to draw samples from a \emph{target distribution} when directly sampling from it is difficult and only a function that is proportional to its density function is available \cite{brooks2011handbook}. These algorithms are particularly useful in drawing a set of samples to approximate the posterior distribution of a random variable in Bayesian models.
In the context of our work, MCMC algorithms allow us to acquire a set of model parameters that are likely to be close to the retrained model parameters $\bm{\theta}_{\mcl{D}_r}$ without the knowledge of the erased dataset $\mcl{D}_e$ (or $\mcl{D}_r$), as explained later in Section~\ref{sec:candidateset}.

Consider the ML model in the previous section that is parameterized with parameters $\bm{\theta}$. Given the likelihood of the training dataset $p(\mcl{D}|\bm{\theta})$ and the prior distribution $p(\bm{\theta})$, the posterior distribution of $\bm{\theta}$ given the training data $\mcl{D}$ is obtained with the Bayes' rule:
\begin{align}
    p(\bm{\theta}|\mcl{D}) = p(\mcl{D}|\bm{\theta})p(\bm{\theta}) / p(\mcl{D})
    \propto p(\mcl{D}|\bm{\theta})p(\bm{\theta})\ .\label{eq:proptopost}
\end{align}
While the prior distribution $p(\bm{\theta})$ and the likelihood $p(\mcl{D}|\bm{\theta})$ are often available from the model, it is difficult to evaluate the marginal likelihood (or the evidence) $p(\mcl{D})~=~\int p(\mathcal{D}|\bm{\theta})p(\bm{\theta})\ \text{d}\bm{\theta}$, e.g., when $p(\bm{\theta})$ is not a conjugate prior for the likelihood function $p(\mcl{D}|\bm{\theta})$. On the other hand, MCMC algorithms are still able to generate the samples of this target distribution $p(\bm{\theta}|\mcl{D})$ using $p(\mcl{D}|\bm{\theta})p(\bm{\theta})$ (which does not involve $p(\mcl{D})$) as this function is proportional to the posterior distribution density $p(\bm{\theta}|\mcl{D})$, as seen in Eq.~\eqref{eq:proptopost}.

\begin{algorithm}[t]
   \caption{Metropolis-Hastings Algorithm}
\begin{algorithmic}[1]
   \State {\bfseries Input:} $f(\bm{\theta}) = p(\mcl{D}|\bm{\theta}) p(\bm{\theta})$, proposal density $g(\bm{\theta}'|\bm{\theta})$, initial sample $\bm{\theta}_0$, number of samples $M$
   \For{$i=1,2,\dots, M$}
   \State Draw a proposed sample $\bm{\theta}'$ from $g(\bm{\theta}|\bm{\theta}_{i-1})$
   \State Evaluate the acceptance ratio:
        \begin{align}
        \beta = \frac{f(\bm{\theta}')}{f(\bm{\theta}_{i-1})} =  \frac{p(\mcl{D}|\bm{\theta}') p(\bm{\theta}')}{p(\mcl{D}|\bm{\theta}_{i-1}) p(\bm{\theta}_{i-1})}\ .
        \label{eq:acratio}
        \end{align}
   \State Draw a uniform random number $u \in [0,1]$.
   \If {$u \le \beta$} \Comment{Accept sample with prob. $\min(\beta,1)$}
      \State $\bm{\theta}_i = \bm{\theta}'$
   \Else \Comment{Reject proposed sample}
      \State $\bm{\theta}_i = \bm{\theta}_{i-1}$ \Comment{Reuse previous sample}
   \EndIf
   \EndFor
   \State \Return $\{\bm{\theta}\}_{i=1}^M$
\end{algorithmic}
\label{alg:metropolis}
\end{algorithm}
MCMC algorithms construct a Markov chain whose equilibrium distribution is the target distribution, e.g.,  $p(\bm{\theta}|\mcl{D})$. A classic MCMC method is the Metropolis-Hastings (M-H) algorithm described in Algorithm~\ref{alg:metropolis} \cite{metropolis1953equation,hastings1970monte}. The algorithm constructs a Markov chain starting from an initial sample $\bm{\theta}_0$ that can be selected arbitrarily. The \emph{proposal density} $g(\bm{\theta}'|\bm{\theta})$ is used to draw the next sample given the current sample. It is chosen as a symmetric distribution centered at the current sample such as a Gaussian distribution. Then the next sample is accepted with the probability $\min(\beta,1)$ where $\beta$ is the acceptance ratio in Eq.~\eqref{eq:acratio}. 
As the target distribution is only approximated well by the equilibrium distribution of the Markov chain, several initial samples are often discarded (which are called \emph{burn-in} samples).

Nonetheless, the M-H algorithm requires a large number of samples to approximate the target density well if the acceptance ratio is low. There have been several MCMC methods that improve the sampling efficiency for a target density of a moderately high dimensional random variable by utilizing the Hamiltonian dynamics such as the \emph{Hamiltonian Monte Carlo} (HMC) \cite{betancourt2017conceptual,neal2011} and the no-u-turn sampler \cite{hoffman2014no}. There are also methods scalable to large datasets by relying on the stochastic gradients \cite{welling2011bayesian,chen2014stochastic,zhang2019cyclical} and a symmetric splitting integration scheme for HMC \cite{cobb2020scaling}. It is noted that we do not focus on addressing these issues of MCMC in this work. The main focus of our work is to design a MU algorithm given a set of MCMC samples that approximate the posterior distribution well.

While MCMC algorithms have been popular methods to draw samples from a target distribution in many ML applications, there are also certain cases such as those in our \acron{} approach where we would like to obtain a set of samples representing a distribution from another set of samples representing a similar yet different distribution, by assigning weights to these samples. This technique is called \emph{importance sampling}.
In particular, we will employ importance sampling to transform the set of MCMC samples representing the distribution  $p(\bm{\theta}|\mcl{D})$ to an approximation of the distribution $p(\bm{\theta}|\mcl{D}_r)$ in Section~\ref{sec:unlearnwithcandidate} and Section~\ref{sec:unlearnwithenlarge}.

\subsection{Importance Sampling}
\label{sec:importancesampling}
Suppose we are interested in estimating the mean of a random variable $f(\mbf{x})$ where $\mbf{x}$ follows a distribution specified by $p(\mbf{x})$. It can be estimated by:
\begin{equation*}
    \mbb{E}[f(\mbf{x})] \triangleq \int f(\mbf{x}) p(\mbf{x})\ \text{d}\mbf{x} \approx \frac{1}{|\mcl{X}|} \sum_{\mbf{x} \in \mcl{X}} f(\mbf{x})
\end{equation*}
where $\mcl{X}$ is a set of samples of $\mbf{x}$ drawn from the distribution specified by $p(\mbf{x})$.

\begin{figure}
    \centering
    \begin{tabular}{@{}c@{}}
    \includegraphics[width=0.38\textwidth]{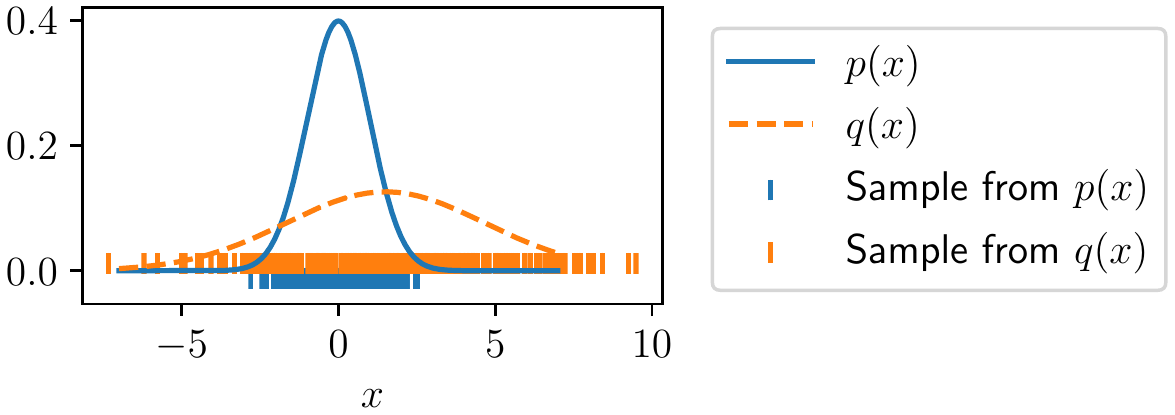}
    \\
    (a) Samples of $p(x)$ and $q(x)$.
    \vspace{2mm}
    \\
    \includegraphics[width=0.32\textwidth]{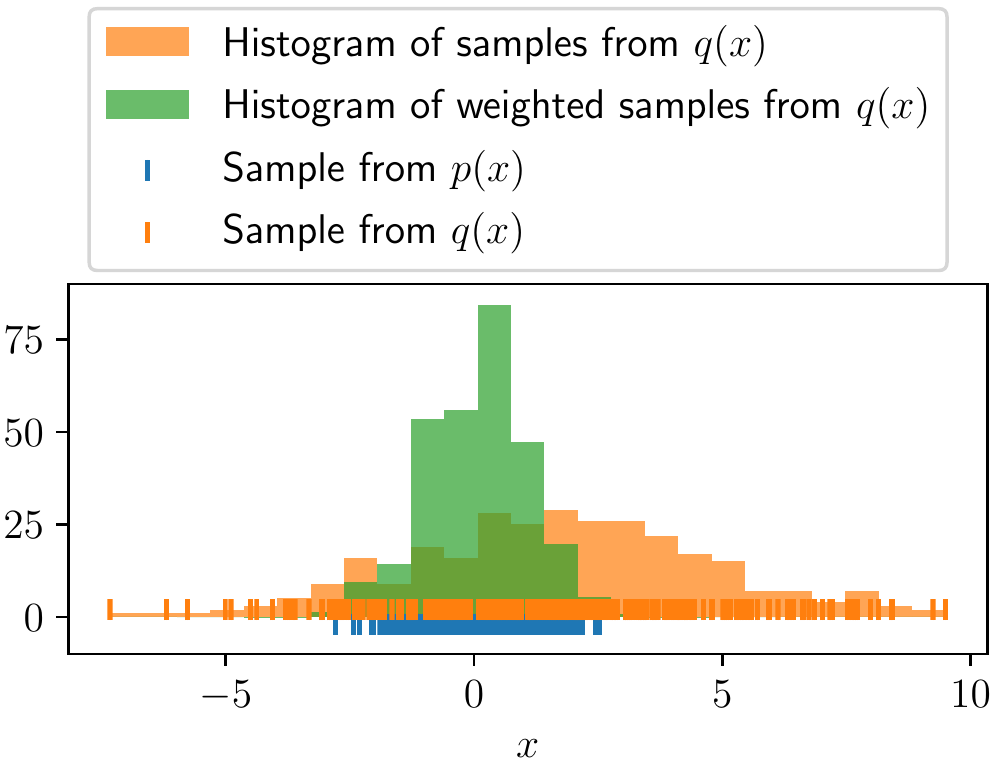}
    \\
    \shortstack{(b) The weighted samples of $q(x)$
    with importance sampling.}
    \\
    \end{tabular}
    \caption{Plots of the weighted samples representing $p(x)$ obtained from the samples of $q(x)$ with importance sampling.}
    \label{fig:demoimpsam}
    \vspace{-0.2cm}
\end{figure}

However, the problem arises when it is difficult to draw samples from the distribution specified by $p(\mbf{x})$. In such case, if there exists a distribution specified by $q(\mbf{x})$ from which samples of $\mbf{x}$ can be drawn easily, then \emph{importance sampling} is a popular technique to estimate $\mbb{E}[f(\mbf{x})]$ using a set $\mcl{X}'$ of samples drawn from the distribution specified by $q(\mbf{x})$:\footnote{Importance sampling requires the condition if $p(\mbf{x}) > 0$ then $q(\mbf{x}) > 0$ so that the density ratio $p(\mbf{x}) / q(\mbf{x})$ is defined for all $\mbf{x}$.}
\begin{equation}
    \mbb{E}[f(\mbf{x})] 
        = \int q(\mbf{x}) \frac{p(\mbf{x})}{q(\mbf{x})} f(\mbf{x})\ \text{d}\mbf{x}
        \approx \frac{1}{|\mcl{X}'|} \sum_{\mbf{x} \in \mcl{X}'} \frac{p(\mbf{x})}{q(\mbf{x})} f(\mbf{x})\ .\label{eq:impsamp} 
\end{equation}
As an illustration, Figure~\ref{fig:demoimpsam} shows that the samples $x \in \mcl{X}'$ with the weights defined as the density ratio $p(\mbf{x})/q(\mbf{x})$ are able to represent the distribution specified by $p(x)$. The plots of $p(x)$ and $q(x)$ are shown in Figure~\ref{fig:demoimpsam}a. We observe that the two densities are different, so their samples are distributed differently.
Suppose we cannot draw samples from $p(x)$ and we can only draw samples from $q(x)$. 
Using the importance sampling technique, we can obtain a set of weighted samples that represent the distribution specified by $p(x)$ from the set of samples of the distribution specified by $q(x)$. 
In particular, the samples from the distribution specified by $q(x)$ are weighted with the density ratio $p(x) / q(x)$ (in a similar fashion to Eq.~\eqref{eq:impsamp}).
We observe in Figure~\ref{fig:demoimpsam}b that the histogram of the weighted samples obtained from importance sampling can represent the distribution specified by $p(x)$.

\section{MCMC-Based Machine Unlearning}
\label{sec:mcu}

We first define the threat model, and then explicitly and formally define the problem of machine unlearning. Common notations used in this paper are listed in Table~\ref{tab:notations}.

{\bf Threat model.} This work considers two classes of threats in ML systems. (a)~A subset of the data used for training is malicious or poisoned by an adversary. This essentially means part of the labels provided as ground truth is wrong. This can happen due to errors in data collection and labeling, but could also be the result of an adversarial attack~\cite{szegedy2013intriguing,goodfellow2014explaining,koh2017understanding,backdoor-poisoning-malware-2021}; both are relevant as they affect model performance, in terms of classification accuracy. (b) The second class of threats comes from how an ML model needs to be maintained to respect user privacy. A user exercising her right to withdraw the use of her data for any purposes previously agreed to, requires the ML model to erase the data lineage corresponding to the user.
To examine whether a model has unlearned a given user's data lineage, we need to evaluate the performance of the unlearned model on the user's data. Contrary to the traditional objective of achieving higher accuracy, note that the unlearned model can have lower classification accuracy of the erased dataset.

{\bf Problem Definition.} 
The problem is to remove the effects of a set $\mcl{D}_e$ of unwanted data 
(also referred to as {\em erased dataset}) 
in a model trained on $\mcl{D} \supset \mcl{D}_e$, by finding an approximation of the model parameters $\bm{\theta}_{\mcl{D}_r}$ without involving the costly procedure of retraining the model on $\mcl{D}_r \triangleq \mcl{D} \setminus \mcl{D}_e$.
We make the assumption that the size of the erased dataset $\mcl{D}_e$ is relatively small compared with that of the training dataset $\mcl{D}$. 

{\bf{Overview of \acron{}.}} We approach the problem by constructing a \emph{candidate set} $\bm{\Theta}$ of the model parameters that is close to the parameters $\bm{\theta}_{\mcl{D}_r}$ without the knowledge of $\mcl{D}_e$ (or $\mcl{D}_r$) (Section~\ref{sec:candidateset}). While the construction of $\bm{\Theta}$ involves running an MCMC algorithm, it is pre-computed before the unlearning procedure because it does not require the knowledge of $\mcl{D}_e$. Therefore, given a dataset $\mcl{D}_e$ to be erased from the trained model, the pre-computed $\bm{\Theta}$ (and its pre-computed auxiliary values) can be used to unlearn the model quickly (Section~\ref{sec:unlearnwithcandidate}). 
We further enlarge the candidate set so that the chance it is closer to the parameters $\bm{\theta}_{\mcl{D}_r}$ increases; this is achieved through the introduction of the \emph{enlarged candidate set} by flattening the posterior distribution $p(\bm{\theta}|\mcl{D})$ (Section~\ref{sec:enlargecandidateset}). Due to the difference between the flattened distribution and $p(\bm{\theta}|\mcl{D})$, importance sampling (Section~\ref{sec:importancesampling}) is utilized to perform unlearning with this enlarged candidate set (Section~\ref{sec:unlearnwithenlarge}).
\begin{table}
    \centering
    \caption{Table of notations.}
    \begin{tabular}{ll}
         \toprule
         Notation & Definition\\
         \midrule
         $\mcl{D}$ & The training dataset\\
         $\mcl{D}_e$ & The erased dataset ($\mcl{D}_e \subset \mcl{D}$)\\
         $\mcl{D}_r$ & The remaining dataset (i.e., $\mcl{D}_r \triangleq \mcl{D} \setminus \mcl{D}_e$)\\
         \midrule
         $\bm{\theta}_{\mcl{D}}$ & The parameters of the model trained on $\mcl{D}$\\
         $\bm{\theta}_{\mcl{D}_r}$ & The parameters of the model (re)trained on $\mcl{D}_r$\\
         \midrule
         $\bm{\Theta}$ & The candidate set\\
         $w(\bm{\theta})$ & The weight of a candidate $\bm{\theta} \in \bm{\Theta}$\\
         \midrule
         \makecell[l]{$\tilde{p}(\bm{\theta}|\mcl{D};\alpha)$\\ \\
         } & \makecell[l]{A flattened distribution of $p(\bm{\theta};\mcl{D})$, defined as\\ 
         $\tilde{p}(\bm{\theta}|\mcl{D};\alpha) \propto \left(p(\bm{\theta})p(\mcl{D}|\bm{\theta})\right)^\alpha$ for $\alpha \in (0,1]$}\\
         $\widetilde{\bm{\Theta}}(\alpha)$ & The enlarged candidate set at the scale $\alpha$\\
         \makecell[l]{$\tilde{\beta}(\alpha)$\\ \\ \\} & \makecell[l]{The acceptance ratio in the M-H algorithm\\
         when applied to drawing samples\\
         from $\tilde{p}(\bm{\theta}|\mcl{D};\alpha)$}\\
         $\tilde{w}(\bm{\theta})$ & The weight of a candidate $\bm{\theta} \in \widetilde{\bm{\Theta}}(\alpha)$\\
         \bottomrule
    \end{tabular}
    \label{tab:notations}
\end{table}

\subsection{Candidate Set of Unlearned Parameters}
\label{sec:candidateset}
Let us consider a discrete set $\bm{\Theta}$ which we refer to as the \emph{candidate set} of unlearned model parameters. This set $\bm{\Theta}$ is constructed without the knowledge of the erased dataset $\mcl{D}_e$. 
The intention in the design of $\bm{\Theta}$ is that, given an erased dataset $\mcl{D}_e$, the parameters $\bm{\theta}_{\mcl{D}_r}$ (unknown when $\bm{\Theta}$ is constructed) are close to a candidate in $\bm{\Theta}$.

As $\mcl{D}_e$ is unknown during the construction of $\bm{\Theta}$, we rely on the assumption that $\mcl{D}_e$ is relatively small compared with $\mcl{D}$. Therefore, after unlearning the model (with parameters $\bm{\theta}_{\mcl{D}}$) trained on $\mcl{D}$ from $\mcl{D}_e$, we assume that the obtained model parameters $\bm{\theta}_{\mcl{D}_r}$ do not differ much from $\bm{\theta}_{\mcl{D}}$. Furthermore, as $\bm{\theta}_{\mcl{D}}$ is the MAP estimate of the model parameters given $\mcl{D}$, $\bm{\theta}_{\mcl{D}}$ is the mode of the posterior distribution of $\bm{\theta}$ given $\mcl{D}$ (see Section~\ref{sec:machineunlearning}).
As a result, from our assumption that the model parameters $\bm{\theta}_{\mcl{D}_r}$ do not differ much from $\bm{\theta}_{\mcl{D}}$, $\bm{\theta}_{\mcl{D}_r}$ does not differ much from the mode of the posterior distribution of $\bm{\theta}$ given $\mcl{D}$. In other words, the posterior density $p(\bm{\theta}_{\mcl{D}_r}|\mcl{D})$ is sufficiently large.

Hence, in order to construct the candidate set $\bm{\Theta}$ that is likely to contain the parameters of the model unlearned from $\mcl{D}_e$, we would like to construct $\bm{\Theta}$ as the set of model parameters $\bm{\theta}$ with high posterior probability densities $p(\bm{\theta}|\mcl{D})$. To this end, we propose to use MCMC methods to draw samples from the posterior distribution of $\bm{\theta}$ given $\mcl{D}$, e.g., applying Algorithm~\ref{alg:metropolis} with $f(\bm{\theta}) = p(\mcl{D}|\bm{\theta})p(\bm{\theta})$. Therefore, the candidate set $\bm{\Theta}$ is a set of samples drawn the distribution $p(\bm{\theta}|\mcl{D})$. 

\vspace{-2mm}
\subsection{MU with Candidate Set} 
\label{sec:unlearnwithcandidate}

While the constructed candidate set $\bm{\Theta}$ is likely to contain the unlearned model parameters if they do not differ much from $\bm{\theta}_{\mcl{D}}$, we still need to construct the unlearned model parameters given this candidate set $\bm{\Theta}$ and an erased dataset~$\mcl{D}_e$. 

Together with the candidate set $\bm{\Theta}$, we also store the values $h(\bm{\theta}) \triangleq \log p(\mcl{D}|\bm{\theta}) + \log p(\bm{\theta})$ for all candidates $\bm{\theta} \in \bm{\Theta}$. MCMC algorithms often require this value to evaluate the acceptance ratio (e.g., in Algorithm~\ref{alg:metropolis}), so it does not incur additional computation to evaluate $\log p(\mcl{D}|\bm{\theta}) + \log p(\bm{\theta})$. As a result, both the candidate set $\bm{\Theta}$ and these auxiliary values $h(\bm{\theta})$ can be pre-computed before the unlearning happens.

When there is a request to unlearn the trained model from an erased dataset $\mcl{D}_e$, we make use of $h(\bm{\theta}) \triangleq \log p(\mcl{D}|\bm{\theta}) + \log p(\bm{\theta})$ to efficiently evaluate the following value:
\begin{align}
&\log p(\bm{\theta}|\mcl{D}_r) 
    = \log p(\mcl{D}_r|\bm{\theta}) + \log p(\bm{\theta}) - \log p(\mcl{D}_r)\nonumber\\
    &= h(\bm{\theta}) - \log p(\mcl{D}_e|\bm{\theta}) - \log p(\mcl{D}_r)
    = g(\bm{\theta},\mcl{D}_e) - \log p(\mcl{D}_r)
    \label{eq:postremain}
\end{align}
for $\bm{\theta} \in \bm{\Theta}$, where $g(\bm{\theta},\mcl{D}_e) \triangleq h(\bm{\theta}) - \log p(\mcl{D}_e|\bm{\theta})$  and we use the assumption that the data are conditionally independent given the model parameters (Eq.~\eqref{eq:condindep}) and $(\mcl{D}_r, \mcl{D}_e)$ is a partition of $\mcl{D}$, i.e.,
\begin{align*}
\log p(\mcl{D}|\bm{\theta}) = \log p(\mcl{D}_r \cup \mcl{D}_e|\bm{\theta})
    = \log p(\mcl{D}_r |\bm{\theta}) + \log p(\mcl{D}_e|\bm{\theta})\ .
\end{align*}
In Eq.~\eqref{eq:postremain}, $\log p(\mcl{D}_r)$ is independent of $\bm{\theta}$, so it is treated as a constant.
Besides, we note that i) we do not use $\mcl{D}_r$ in the evaluation of $g(\bm{\theta}, \mcl{D}_e)$ in Eq.~\eqref{eq:postremain}; and ii) given the stored value $h(\bm{\theta}) \triangleq \log p(\mcl{D}|\bm{\theta}) + \log p(\bm{\theta})$ and $\mcl{D}_e$, the evaluation of $g(\bm{\theta}, \mcl{D}_e)$ is efficient as we assume that $\mcl{D}_e$ is small.

We recall that the ultimate goal of MU is to obtain the parameters that is close to the parameters $\bm{\theta}_{\mcl{D}_r}$ of the model retrained on $\mcl{D}_r$, i.e., the parameters that maximizes the log posterior probability $\log p(\bm{\theta}|\mcl{D}_r)$. 
Note that the candidate set $\bm{\Theta}$ is likely to contain such parameters as explained in Section~\ref{sec:candidateset}, and $g(\bm{\theta},\mcl{D}_e)$ differs with $\log p(\bm{\theta}|\mcl{D}_r)$ by a constant $\log p(\mcl{D}_r)$ for all $\bm{\theta} \in \bm{\Theta}$.
Therefore, we can choose the candidate in $\bm{\Theta}$ with the maximum posterior probability, given the remaining dataset by choosing the candidate $\bm{\theta} \in \bm{\Theta}$ with the largest value of $g(\bm{\theta}, \mcl{D}_e)$. 

More importantly, we are also able to obtain an approximate distribution of the posterior distribution given the remaining dataset (i.e., $p(\bm{\theta}|\mcl{D}_r)$) based on importance sampling.
Recall that $\bm{\Theta}$ is constructed as MCMC samples from the posterior distribution $p(\bm{\theta}|\mcl{D})$, so we can assign weight $w(\bm{\theta})$ to each candidate $\bm{\theta}$ in $\bm{\Theta}$ as follows
\begin{align}
    w(\bm{\theta}) =  \frac{p(\bm{\theta}|\mcl{D}_r)}{p(\bm{\theta}|\mcl{D})} 
        = \frac{p(\mcl{D}_r|\bm{\theta}) p(\bm{\theta})}{p(\mcl{D}|\bm{\theta}) p(\bm{\theta})} \frac{p(\mcl{D})}{p(\mcl{D}_r)}
	= \frac{e^{g(\bm{\theta},\mcl{D}_e)}}{e^{h(\bm{\theta})}} \frac{p(\mcl{D})}{p(\mcl{D}_r)}\label{eq:weight1}
\end{align}
where $p(\mcl{D})/p(\mcl{D}_r)$ is independent of $\bm{\theta}$, so it disappears after we normalize the weights for all $\bm{\theta} \in \bm{\Theta}$. As illustrated in Figure~\ref{fig:demoimpsam}, we can use this weighted set $\bm{\Theta}$ to approximate the posterior distribution $p(\bm{\theta}|\mcl{D}_r)$. Therefore, we can also use the weighted average of $\bm{\Theta}$ as the unlearned model parameters.

\subsection{Enlarged Candidate Set}
\label{sec:enlargecandidateset} 

\begin{figure*}[t]
    \centering
    \begin{tabular}{@{}c@{}c@{}c@{}c@{}c@{}}
        \includegraphics[width=0.2\textwidth]{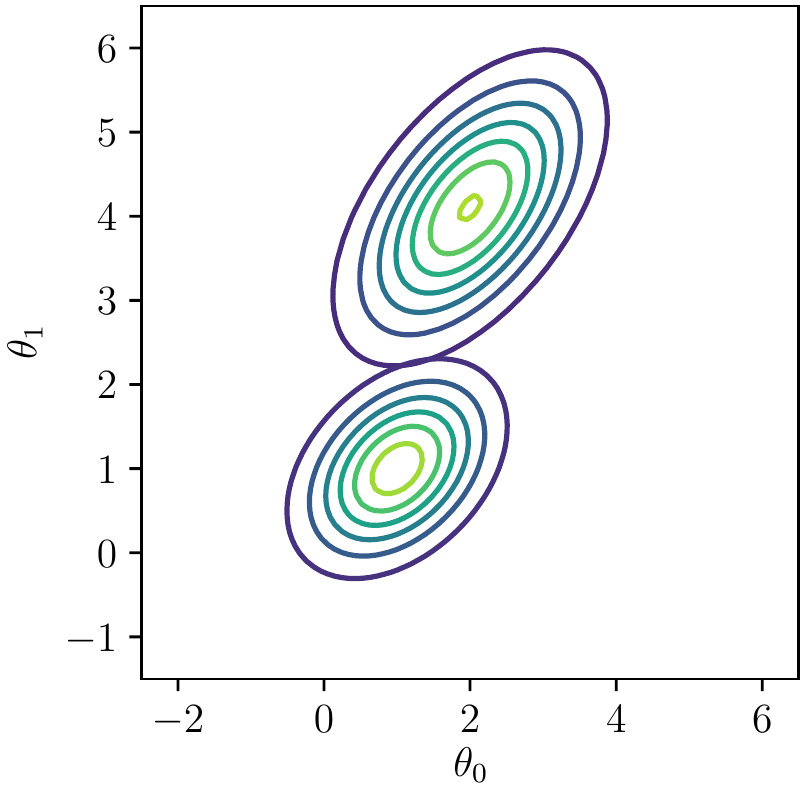}
        &
        \includegraphics[width=0.2\textwidth]{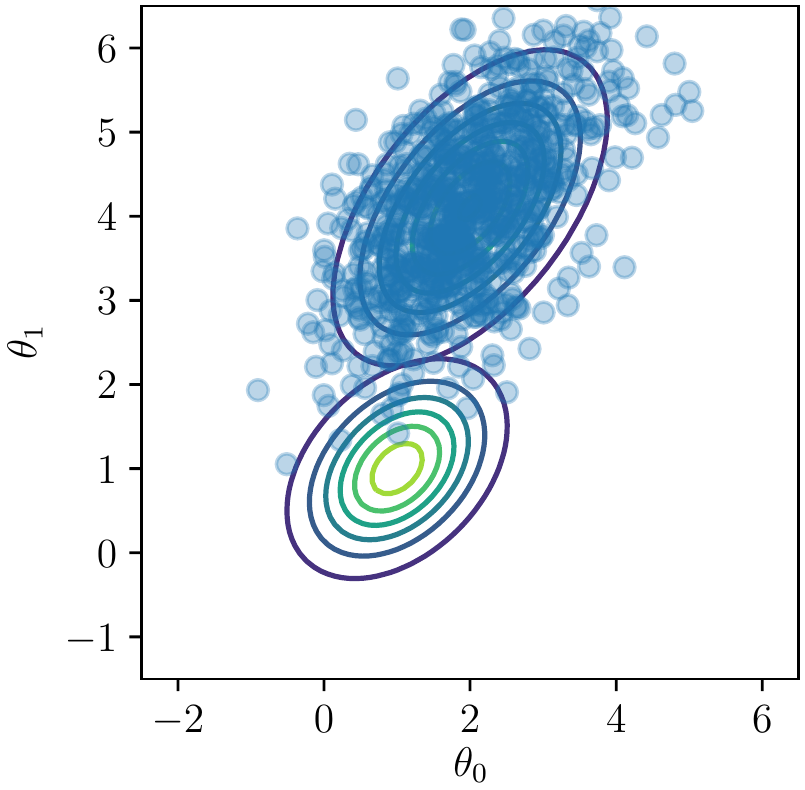} 
        &
        \includegraphics[width=0.2\textwidth]{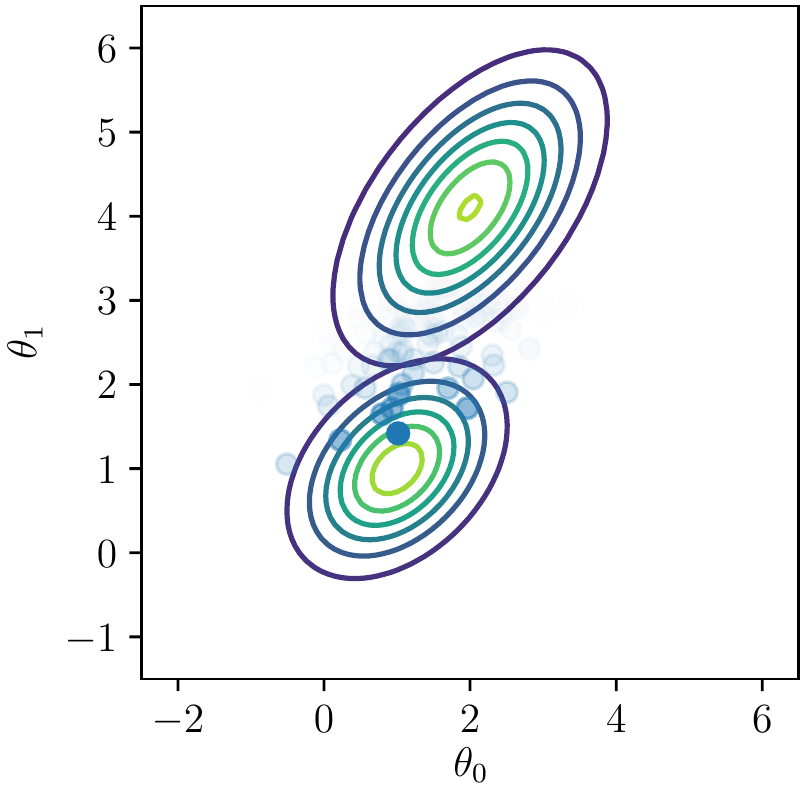}
        &
        \includegraphics[width=0.2\textwidth]{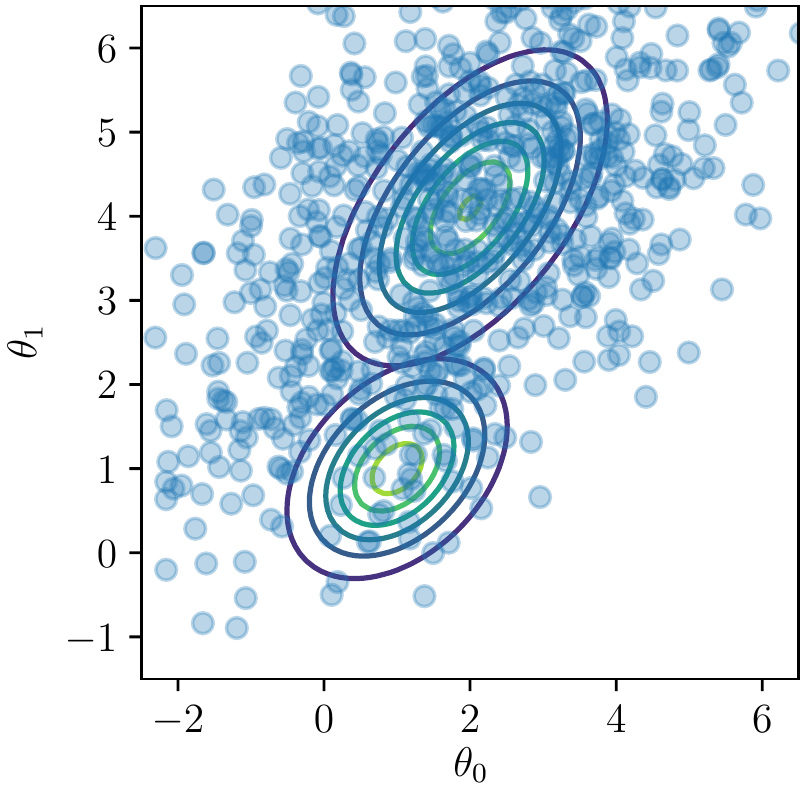} 
        &
        \includegraphics[width=0.2\textwidth]{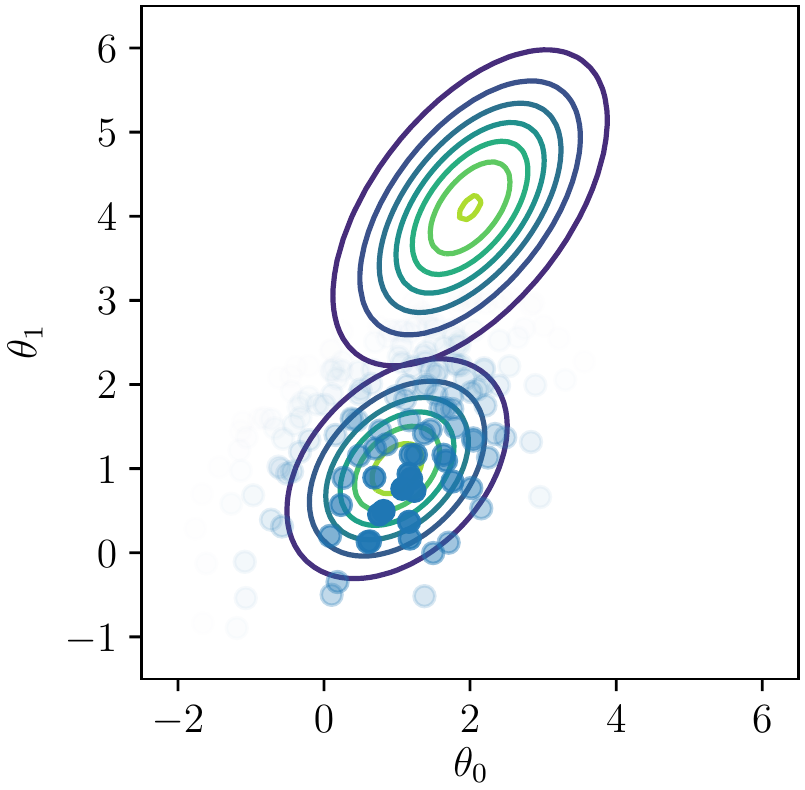}\\
        (a) & (b) & (c) & (d) & (e)\vspace{-3mm}
    \end{tabular}
    \caption{Plots of a hypothetical MU experiment with two parameters $\theta_0$ and $\theta_1$. (a) shows $p(\bm{\theta}|\mcl{D})$ and $p(\bm{\theta}|\mcl{D}_r)$ as Gaussian distributions centered at $(2,4)$ and $(1,1)$, respectively. (b) shows the samples drawn from $p(\bm{\theta}|\mcl{D})$. (c) shows the weighted samples from those in (b) to represent $p(\bm{\theta}|\mcl{D}_r)$. (d) shows the samples drawn from a flattened distribution $\tilde{p}(\bm{\theta}|\mcl{D};\alpha)$. (e) shows the weighted samples from those in (d) to represent $p(\bm{\theta}|\mcl{D}_r)$.}
    \label{fig:demoflatten}
    \vspace{-0.2cm}
\end{figure*}

\begin{remark}
\label{remark:enlarge}
The main rationale behind the construction of the candidate set $\bm{\Theta}$ is that if the erased dataset $\mcl{D}_e$ is small compared with $\mcl{D}$, the retrained model parameters $\bm{\theta}_{\mcl{D}_r}$ should have a sufficient high posterior probability density $p(\bm{\theta}_{\mcl{D}_r}|\mcl{D})$. However, in practice, by only performing MCMC sampling for the posterior distribution $p(\bm{\theta}|\mcl{D})$, we may not be able to obtain a sample that is close to $\bm{\theta}_{\mcl{D}_r}$. Figure~\ref{fig:demoflatten}a shows a hypothetical scenario: the contour plot of a Gaussian distribution centered at $(2,4)$ represents the posterior distribution $p(\bm{\theta}|\mcl{D})$ of the model parameters given $\mcl{D}$ while the contour plot of a Gaussian distribution centered at $(1,1)$ represents the posterior distribution $p(\bm{\theta}|\mcl{D}_r)$ of the model parameters given $\mcl{D}_r$.
As the change in the posterior distribution of the model parameters after unlearning (i.e., $p(\bm{\theta}|\mcl{D}_r)$ vs. $p(\bm{\theta}|\mcl{D})$) is sufficiently large, we observe that samples drawn from $p(\bm{\theta}|\mcl{D})$ (which constitute our candidate set $\bm{\Theta}$) do not overlap with the region of high posterior probability $p(\bm{\theta}|\mcl{D}_r)$ in Figure~\ref{fig:demoflatten}b.
In fact, by plotting these samples/candidates with the weights from importance sampling in Eq.~\eqref{eq:weight1} in Figure~\ref{fig:demoflatten}c, we observe that they do not represent the distribution $p(\bm{\theta}|\mcl{D}_r)$ well due to the lack of samples/candidates in the region of high posterior probability $p(\bm{\theta}|\mcl{D}_r)$.
\end{remark}

Consequently, based on Remark~\ref{remark:enlarge}, we would like to enhance the capability of our approach to address the above issue in this section. We take the approach of enlarging the region where $\bm{\Theta}$ is constructed. It is done by sampling the candidate set $\bm{\Theta}$ from a \emph{flattened distribution} of $p(\bm{\theta}|\mcl{D})$, instead of from $p(\bm{\theta}|\mcl{D})$. The flattened distribution is defined as follows.

{\bf{Flattened distribution.}} Consider $p(\bm{\theta}|\mcl{D}) \propto p(\bm{\theta}) p(\mcl{D}|\bm{\theta})$, we construct its flattened distribution as  $\tilde{p}(\bm{\theta}|\mcl{D}; \alpha) \propto \left(p(\bm{\theta}) p(\mcl{D}|\bm{\theta})\right)^\alpha$ for $\alpha \in (0,1]$. While the exact density $\tilde{p}(\bm{\theta}|\mcl{D}; \alpha)$ of this flattened distribution is unknown, its proportional value $\left(p(\bm{\theta}) p(\mcl{D}|\bm{\theta})\right)^\alpha$ is sufficient for us to draw its samples with MCMC algorithms (e.g., Algorithm~\ref{alg:metropolis}). We call the candidate set constructed from this flattened distribution the \emph{enlarged candidate set} at the scale $\alpha$, denoted as $\widetilde{\bm{\Theta}}(\alpha)$. When $\alpha = 1$, $\tilde{p}(\bm{\theta}|\mcl{D};\alpha=1) = p(\bm{\theta}|\mcl{D})$ and $\widetilde{\bm{\Theta}}(1) = \bm{\Theta}$.

\begin{figure}[t]
    \centering
    \includegraphics[width=0.4\textwidth]{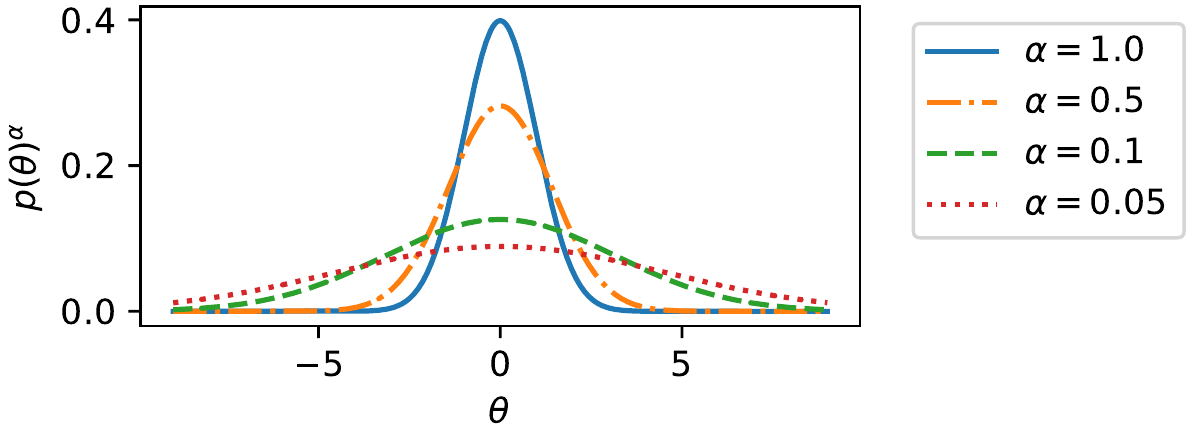}\vspace{-0.3cm}
    \caption{Distributions obtained by flattening a distribution with the density $p(\bm{\theta})$ through raising the density to a power $\alpha \in \{1.0, 0.5, 0.1, 0.05\}$ ($\alpha = 1$ does not flatten the distribution).}
    \vspace{-0.2cm}
    \label{fig:flattengauss1d}
\end{figure}

The reason that raising the probability density of a distribution to a power $\alpha  \in (0,1]$ has the effect of flattening the distribution can be explained through the samples drawn from the flattened distribution with MCMC algorithms. Let us denote the acceptance ratio using the probability density raised to a power $\alpha \in (0,1]$ as $\min(\tilde{\beta}(\alpha), 1)$ where $\tilde{\beta}(\alpha)$ is defined in the same manner as  Eq.~\eqref{eq:acratio}: 
\begin{align*}
    \tilde{\beta}(\alpha) = \left(p(\mcl{D}|\bm{\theta}')p(\bm{\theta}')\right)^\alpha / \left(p(\mcl{D}|\bm{\theta}_{i-1})p(\bm{\theta}_{i-1})\right)^\alpha\ .
\end{align*}
We can express this acceptance ratio as a function of another acceptance ratio $\tilde{\beta}(\alpha')$ (i.e., with respect to the target distribution $\tilde{p}(\bm{\theta}|\mcl{D};\alpha')$ for $\alpha' \in (0,1]$).
\begin{align}
    \tilde{\beta}(\alpha) &= \left(p(\mcl{D}|\bm{\theta}')p(\bm{\theta}')\right)^\alpha / \left(p(\mcl{D}|\bm{\theta}_{i-1})p(\bm{\theta}_{i-1})\right)^\alpha
    = \tilde{\beta}(\alpha')^{\alpha / \alpha'}\nonumber\ .
\end{align}
We observe that $\tilde{\beta}$ is a decreasing function for $\alpha \in (0,1)$ if its codomain is restricted to $(0,1)$. For example, $\tilde{\beta}(\alpha) \ge \tilde{\beta}(\alpha')$ if $\alpha \le \alpha'$ and $\tilde{\beta}(\alpha) \in (0,1)$. Therefore, 
\begin{align}
    \min(\tilde{\beta}(\alpha), 1) \ge \min(\tilde{\beta}(\alpha'), 1) \text{ if } \alpha \le \alpha'\nonumber
\end{align}
which means that the acceptance ratio in the M-H algorithm increases when $\alpha$ decreases. As the acceptance ratio increases, the samples drawn from the M-H algorithm cover a larger region, which is illustrated in Figure~\ref{fig:flattengauss1d}.
A special case is when $\alpha \le \alpha' = 1$, $\min(\tilde{\beta}(\alpha), 1) \ge \min(\tilde{\beta}(1), 1) = \min(\beta, 1)$. Thus, the samples drawn from $\tilde{p}(\bm{\theta}|\mcl{D};\alpha)$ for $\alpha \in (0,1)$ using MCMC algorithms cover a larger region of the domain (i.e., flattened) than those samples drawn from $p(\bm{\theta}|\mcl{D})$. 

Flattening the target distribution has also been investigated in the MCMC literature \cite{robert2018accelerating}. It is used to improve the exploration of the MCMC methods, e.g., to discover different modes of the target distribution. This is different from our purpose which is to enlarge the candidate set such that it is close to the unlearned model parameters.

\subsection{MU with Enlarged Candidate Set}
\label{sec:unlearnwithenlarge}

Similar to Section~\ref{sec:unlearnwithcandidate}, to construct the unlearned model parameters given the enlarged candidate set $\widetilde{\bm{\Theta}}(\alpha)$ and an erased dataset $\mcl{D}_e$, we store the value $h(\bm{\theta}) \triangleq~\log~p(\mcl{D}|\bm{\theta})~+~\log p(\bm{\theta})$ for all candidates $\bm{\theta}$ in the enlarged candidate set $\widetilde{\bm{\Theta}}(\alpha)$. Then, we are able to obtain $g(\bm{\theta},\mcl{D}_e) \triangleq h(\bm{\theta}) - \log p(\mcl{D}_e|\bm{\theta})$ that only differs from $\log p(\bm{\theta}|\mcl{D}_r)$ by a constant independent of $\bm{\theta}$ (Eq.~\eqref{eq:postremain}).

However, because the enlarged candidate set $\widetilde{\bm{\Theta}}(\alpha)$ is drawn from the flattened distribution $\tilde{p}(\bm{\theta}|\mcl{D};\alpha)$, the weights in Eq.~\eqref{eq:weight1} cannot be used to construct an approximation to the posterior distribution of $p(\bm{\theta}|\mcl{D}_r)$. Therefore, we make a modification to Eq.~\eqref{eq:weight1} to obtain the weights for each candidate $\bm{\theta} \in \widetilde{\bm{\Theta}}(\alpha)$:
\begin{align}
    &\tilde{w}(\bm{\theta}) =  \frac{p(\bm{\theta}|\mcl{D}_r)}{\tilde{p}(\bm{\theta}|\mcl{D}; \alpha)} 
        = \frac{p(\mcl{D}_r|\bm{\theta}) \tilde{p}(\bm{\theta};\alpha)}{\left(p(\mcl{D}|\bm{\theta}) p(\bm{\theta})\right)^\alpha} \frac{\tilde{p}(\mcl{D};\alpha)}{p(\mcl{D}_r)}
	= \frac{e^{g(\bm{\theta},\mcl{D}_e)}}{e^{\alpha h(\bm{\theta})}} \frac{\tilde{p}(\mcl{D};\alpha)}{p(\mcl{D}_r)}\nonumber
\end{align}
where $\tilde{p}(\mcl{D};\alpha) \triangleq \int \left(p(\mcl{D}|\bm{\theta}) p(\bm{\theta})\right)^\alpha\ \text{d}\bm{\theta}$.
Note that $\tilde{p}(\mcl{D};\alpha)/p(\mcl{D}_r)$ is independent of $\bm{\theta}$, so it disappears after we normalize the weights for all $\bm{\theta} \in \bm{\Theta}$. We illustrate the enlarged candidate set $\widetilde{\bm{\Theta}}(\alpha)$ in Figure~\ref{fig:demoflatten}d and its corresponding weighted candidates in Figure~\ref{fig:demoflatten}e. We can observe that the set of weighted candidates in the enlarged $\widetilde{\bm{\Theta}}(\alpha)$ is able to approximate the posterior distribution $p(\bm{\theta}|\mcl{D}_r)$ much better than the weighted candidates in the candidate set $\bm{\Theta}$ in Figure~\ref{fig:demoflatten}c.
Therefore, we can also use the weighted average of $\widetilde{\bm{\Theta}}(\alpha)$ as the unlearned model parameters.

\section{Explaining the Effect of a Subset of Training Data on Model Prediction}
\label{sec:datainfluence}

While MU has been mainly about removing the effect of a specific subset of training data from the model, we explore a new application of our MU approach \acron{} in explaining the effect of training data on the model prediction. 

Let us consider a scenario that an ML model is trained to detect phishing webpages from data $\mcl{D}$ collected from a number of sources. Let $\mcl{D} = \cup_{i=1}^n \mcl{D}_i$ where $\mcl{D}_i$ is the data labeled by the source $i$. 
We would like to examine if there exists a source that contributes malicious/adversarial training data to the ML model.
A solution is to train a model (A) on the data $\mcl{D}$ collected from all sources. 
To check if a source $i$ contributes malicious/adversarial training data $\mcl{D}_i$, we train a separate model (B\textsubscript{$i$}) on the training data $\mcl{D} \setminus \mcl{D}_i$ (i.e., excluding those from the source $i$).
If the accuracy of the model B\textsubscript{$i$} on a test set (different from the training data) increases significantly compared with that of the model A, then we can say that the data $\mcl{D}_i$ labeled by the source $i$ has a negative effect on our model, i.e., they are potentially malicious/adversarial data. 

As the number of sources contributing to the training data can be large, the amount of time incurred to retrain different models B\textsubscript{$i$} from scratch is prohibitively expensive. Therefore, instead of retraining from scratch, we would like to quickly estimate model B\textsubscript{$i$} using model A and the corresponding data labeled by the source $i$. 
Then, obtaining model B\textsubscript{$i$} in this approach can be viewed as unlearning model A from the data labeled by the source $i$, which is precisely the MU problem in the previous section.

Note that, as the number of sources increases, the approach of retraining becomes more and more expensive. Yet, at the same time, the data labeled by a source is likely to become smaller relative to the whole training dataset. As a result, the approach of MU becomes more practical. 
In Section~\ref{sec:phish},
we demonstrate the effectiveness of \acron{} in explaining the positive impact of subsets of correctly labeled training data and the negative impact of subsets of incorrectly labeled training data with experiments on a phishing webpage detection dataset.

\section{Pitfall: Catastrophic Unlearning}
\label{sec:pitfall}

We describe an important pitfall in MU approaches.

\begin{remark}[Catastrophic unlearning]
\label{remark:catastrophic}
It is noted that we should not unlearn a model from an erased dataset $\mcl{D}_e$ through `reversing' the training procedure by minimizing (instead of maximizing in the learning) the log posterior probability of the erased dataset:
\begin{align}
    \argmin_{\bm{\theta}} \log p(\bm{\theta}|\mcl{D}_e)\ .\nonumber
\end{align}
This is because it may lead to \emph{catastrophic unlearning/forgetting} \cite{du2019lifelong}: the log posterior probability of the remaining dataset $\mcl{D}_r$ is decreased unnecessarily by unlearning. Although the work of \cite{du2019lifelong} mitigates this issue by imposing a lower bound on $\log p(\bm{\theta}|\mcl{D}_e)$ when minimizing $\log p(\bm{\theta}|\mcl{D}_e)$, it does not propose a principled way of setting this lower bound value. Figure~\ref{fig:unlearnmaxprob} shows a synthetic experiments with a linear regression problem.
The dataset consists of tuples $(x,y_x)$. The erased dataset $\mcl{D}_e$ are plotted as orange dots and the remaining dataset $\mcl{D}_r$ are plotted as blue dots. The model prediction of the model trained on $\mcl{D}$ is shown as the dashed green line while that of the model retrained on the remaining dataset $\mcl{D}_r$ is shown as the dashed purple line in the figure. We observe that by removing the orange dots (the erased dataset $\mcl{D}_e$), the purple curve shifts away from the orange dots while it still fits the blue dots (the remaining dataset $\mcl{D}_r$). However, if we unlearn the model from $\mcl{D}_e$ by minimizing the log posterior probability of the erased dataset $\mcl{D}_e$, the unlearned model produces entirely incorrect function values (shown as the red curves) even at the blue dots (the remaining dataset $\mcl{D}_r$). This is the \emph{catastrophic unlearning/forgetting} phenomenon mentioned above. 
Even when the unlearning (i.e., the minimization of $\log p(\bm{\theta}|\mcl{D}_e)$) is reduced by stopping after some number of iterations (e.g., after only $100$ iterations), catastrophic unlearning still happens.

\begin{figure}
    \centering
    \begin{tabular}{@{}c@{}}
         \includegraphics[width=0.35\textwidth]{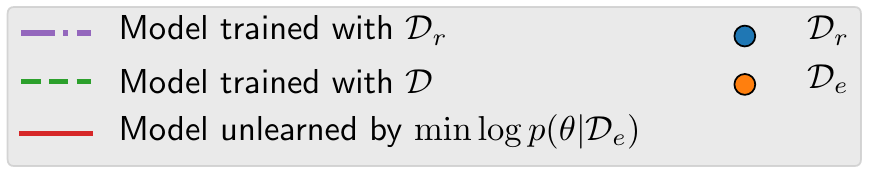}\\
         \includegraphics[width=0.32\textwidth]{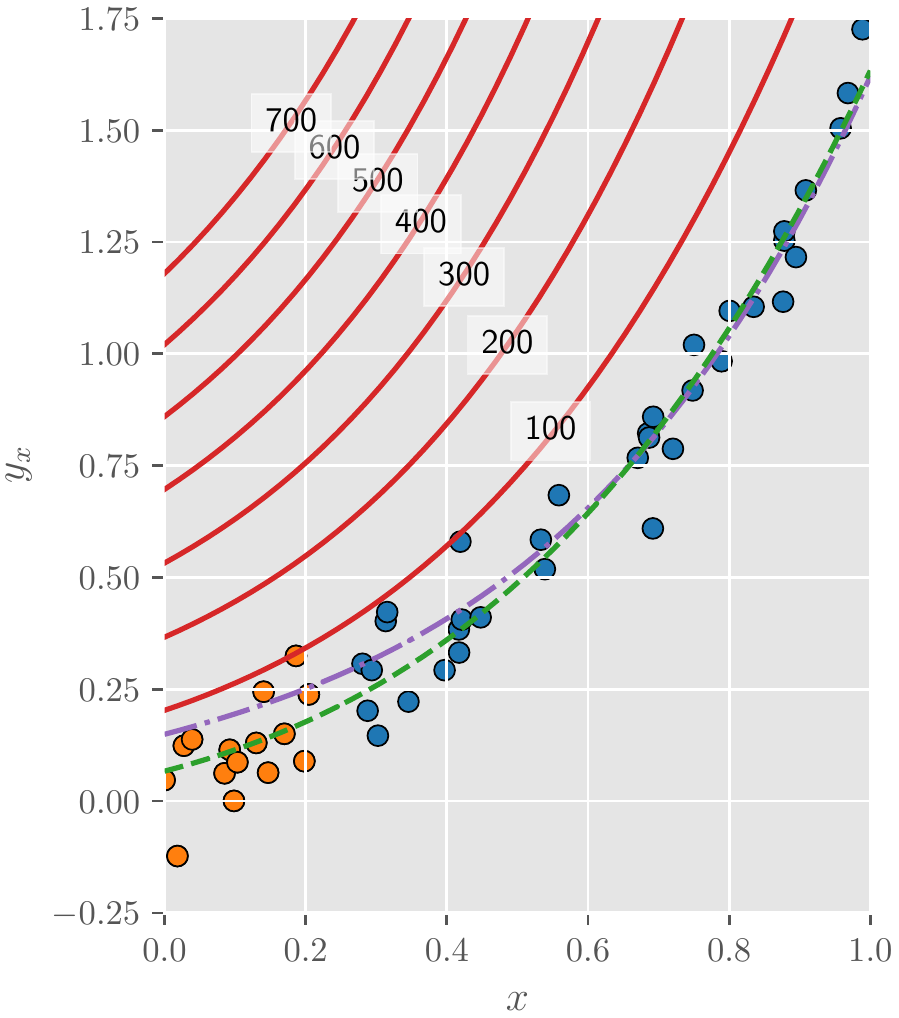}\vspace{-0.3cm}
    \end{tabular}
    \caption{Illustration of the performance of the model unlearned by minimizing $p(\bm{\theta}|\mcl{D}_e)$. The red curves show the predicted function values of the unlearned models with the number of training iterations shown on the curve.}
    \label{fig:unlearnmaxprob}
    \vspace{-0.2cm}
\end{figure}
\end{remark}

As illustrated in the above remark, the approach of minimizing $\log p(\bm{\theta}|\mcl{D}_e)$ is prone to catastrophic unlearning. There also exists approaches based on the influence function \cite{fu2021bayesian,fu22} which performs only a small number of updates using Newton approximation. Thus, it is not prone to catastrophic unlearning as shown in our experiments in Section~\ref{sec:experiments}. However, the method is based on a first-order Taylor approximation, so it is only accurate for a very small change in the model after unlearning.
Our experiments in Section~\ref{sec:experiments} show that our proposed \acron{} approach outperforms this approach even when the unwanted data set is relatively small.

For \acron{}, catastrophic unlearning is mitigated from two angles: i) the (enlarged) candidate set restricts the unlearned model parameters to be within a region of high posterior probability density $p(\bm{\theta}|\mcl{D})$; ii) $g(\bm{\theta},\mcl{D}_e)$ is a good surrogate of the posterior probability density $p(\bm{\theta}|\mcl{D}_r)$. The latter is because $g(\bm{\theta},\mcl{D}_e)$ differs from $p(\bm{\theta}|\mcl{D}_r)$ by a constant (see Eq.~\eqref{eq:postremain}).

\section{Performance evaluation}
\label{sec:experiments}

We empirically illustrate the performance of our proposed \acron{} approach with a binary classification dataset, the Pima Indians diabetes dataset, and the phishing webpage detection dataset. 
The synthetic dataset is deliberately constructed to ease the clutter in the plots (by choosing a sufficiently small training dataset with a low input dimension) while still allowing the effect of unlearning to be easily visualized.
On the other hand, the last two experiments demonstrate the empirical performance of \acron{} in larger real-world datasets and higher input dimensions. 
It is noted that we aim to design a MU algorithm given a set of MCMC samples that estimate the posterior belief of the model parameters well. Besides, it is well-known that performing MCMC sampling for high dimensional distributions is notoriously challenging \cite{barbos17}. Thus, we choose logistic regression models to correctly compare the unlearning performance of \acron{} with that of existing baselines without worrying about the performance of MCMC algorithms.

We study \acron{} in both cases, i)~of using a candidate set $\bm{\Theta}$ and ii)~of using an enlarged candidate set $\widetilde{\bm{\Theta}}(\alpha)$, to highlight the advantage of the latter approach. As explained in 
Section~\ref{sec:enlargecandidateset}, the choice of $\alpha$ should depend on how much the model change after removing the erased dataset. In practice, we suggest setting aside a validation set of erased datasets to tune the value of $\alpha$, i.e., selecting the value of $\alpha$ that the unlearned model obtained from \acron{} with $\widetilde{\bm{\Theta}}(\alpha)$ has the closest performance to the retrained model given an erased dataset in this validation set. 
For our experiments here, we choose the values of $\alpha$ to demonstrate the effect of flattening on \acron{}.

Furthermore, we compare \acron{} with an existing MU algorithm that is also based on MCMC samples, called \emph{Bayesian inference forgetting} (BIF)~\cite{fu2021bayesian,fu22}.
It is worth noting that, BIF utilizes the remaining dataset $\mcl{D}_r$ in the procedure of unlearning, while \acron{} does not. 
Thus, BIF may not be feasible due to the unavailability of $\mcl{D}_r$ or incur substantial overhead as the size of $\mcl{D}_r$ is typically large.
Therefore, we expect BIF to outperform our \acron{} approach in the experiments. However, due to the first-order approximation in BIF, we will observe that \acron{} outperforms BIF empirically in several experiments.

\subsection{Synthetic Binary Classification Dataset}

In this experiment, we train a logistic regression model on a binary classification dataset. We model the log ratio of the probabilities of an input $x$ in the two classes $0$ and $1$ as a polynomial function:
$\log \left(P(y_x = 1) / P(y_x = 0)\right) 
    = \sum_{i=0}^4 a_i x^i$
where $\{a_i\}_{i=0}^4$ are the model parameters. Equivalently, the class probabilities of an input $x$ can be expressed as follows:
\begin{align}
    \textstyle P(y_x = 1) &\textstyle = \exp\left(\sum_{i=0}^4 a_i x^i\right) / \left( 1 + \exp\left(\sum_{i=0}^4 a_i x^i\right)\right)
\end{align}
and $P(y_x = 0) = 1 - P(y_x = 1)$.
The prior distribution of each model parameter is a Gaussian distribution with mean $0$ and variance $5$.

The training dataset $\mcl{D}$ and the erased dataset $\mcl{D}_e$ consist of $50$ and $8$, respectively. Similar to the previous linear regression experiment, we deliberately choose $\mcl{D}_e$ to be a cluster of $8$ data points such that the unlearned model can be easily interpreted. We construct the candidate set and the enlarged candidate set (of the $5$ model parameters $\{a_i\}_{i=0}^4$) using $3000$ MCMC samples.
\begin{figure*}[t]
    \centering
    \begin{tabular}{@{}ccc@{}}
    \multicolumn{3}{c}{\includegraphics[height=0.03\textwidth]{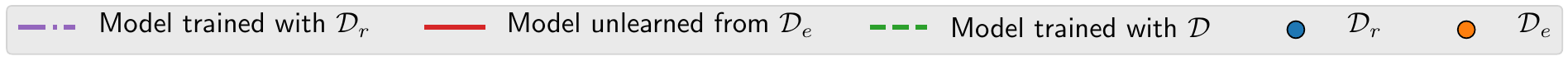}}
    \\
    \includegraphics[width=0.3\textwidth]{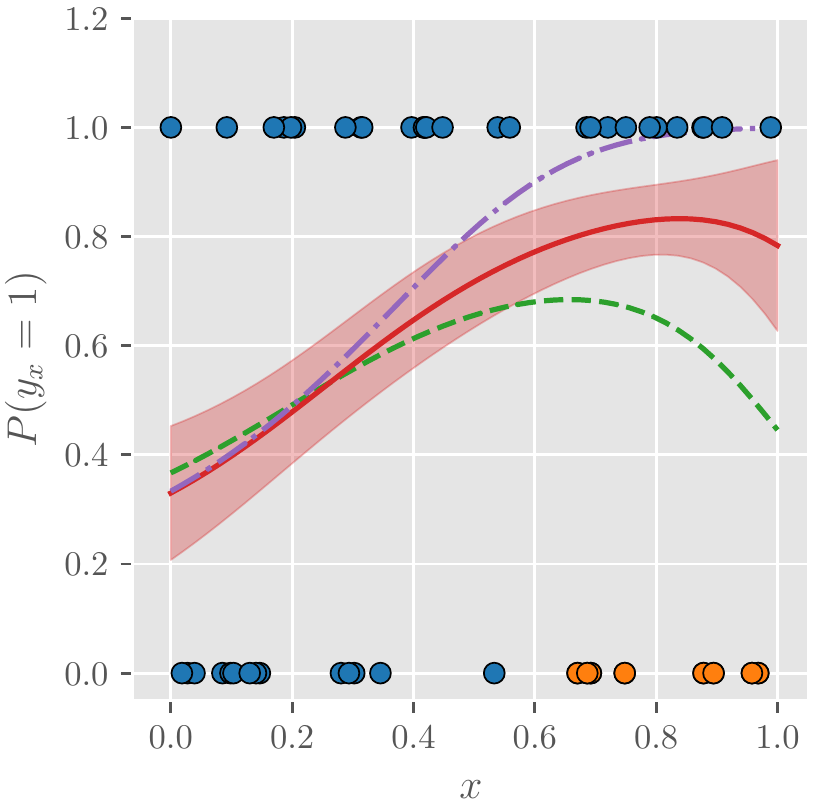}
    &
    \includegraphics[width=0.3\textwidth]{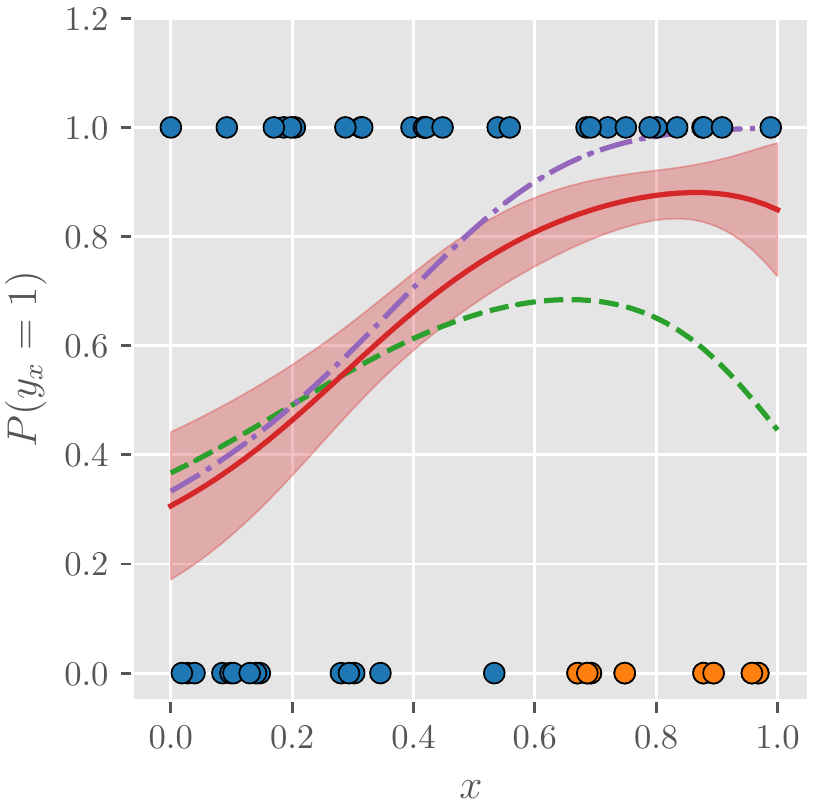}
    &
    \includegraphics[width=0.3\textwidth]{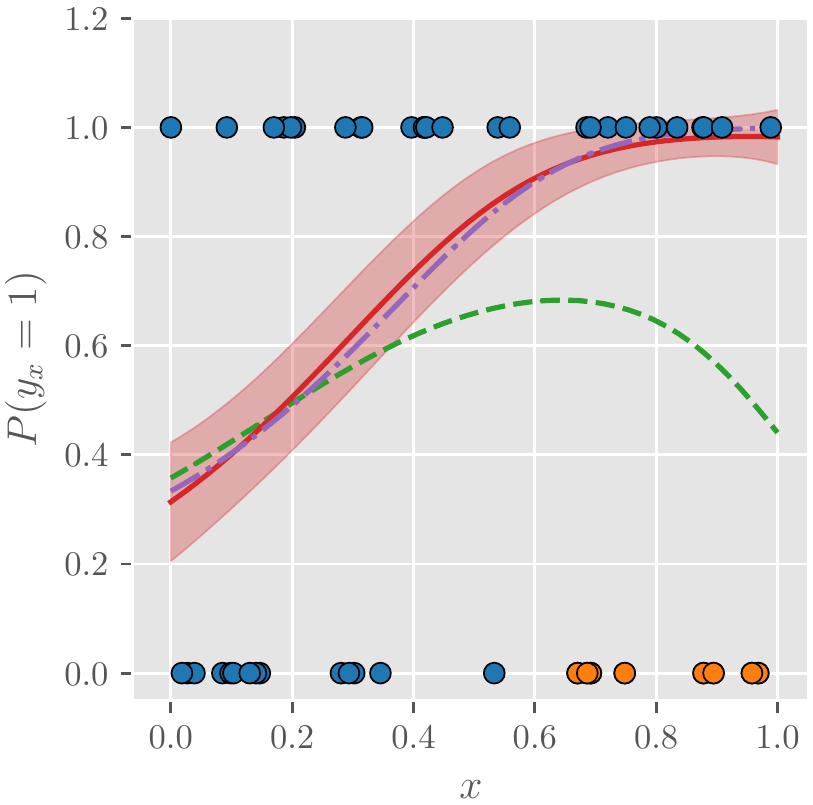}
    \\
    \makecell{(a) BIF.}
    &
    \makecell{(b) \acron{} with $\alpha = 1$.}
    &
    \makecell{(c) \acron{} with $\alpha = 0.1$.}
    \vspace{-0.3cm}
    \end{tabular}
    \caption{Model prediction in experiments on the synthetic binary classification dataset.}
    \label{fig:binaryclass}
\end{figure*}

Without flattening the posterior belief $p(\bm{\theta}|\mcl{D})$ in the construction of the candidate set $\bm{\Theta}$, Figure~\ref{fig:binaryclass}a and~\ref{fig:binaryclass}b show the model predictions of unlearned models with BIF and our \acron{} approach with $\alpha = 1$, respectively.
We observe that the predicted probabilities $P(y_x = 1)$ of these approaches (plotted as red curves) differ from that of the model retrained on $\mcl{D}_r$ (plotted as dashed purple curves), especially in the input region $[0.7,1.0]$ (i.e., the location of the erased dataset $\mcl{D}_e$). However, we also observe that these unlearning approaches can shift the predicted probabilities $P(y_x = 1)$ away from that of the model trained on $\mcl{D}$ (plotted as dashed green curves) and towards that of the model retrained on $\mcl{D}_r$ (plotted as dashed purple curves).

When the enlarged candidate set $\widetilde{\bm{\Theta}}(\alpha)$ is constructed by flattening the posterior belief $p(\bm{\theta}|\mcl{D})$ with $\tilde{p}(\bm{\theta}|\mcl{D};\alpha=0.1)$, the unlearning results are shown in Figure~\ref{fig:binaryclass}c. 
Compared with Figure~\ref{fig:binaryclass}a and~\ref{fig:binaryclass}b for unlearning results with BIF and $\alpha = 1.0$, the predicted $P(y_x=1)$ of the unlearned model in Figure~\ref{fig:binaryclass}c (plotted as a dashed green curve) is closer to that of the model retrained on $\mcl{D}_r$ (plotted as a dashed purple curve). 
The evaluation shows that our proposed \acron{} method with enlarged $\widetilde{\bm{\Theta}}(\alpha)$ outperforms the other two methods in this experiment. Again, \acron{} outperforms BIF which uses the remaining dataset in the unlearning procedure.

\subsection{Pima Indians Diabetes Dataset}

In this experiment, we make use of the Pima Indians diabetes dataset \cite{Dua:2019} to construct the scenario when several patients would like to remove their medical data from a classification model that is trained on their data. There are a total of $768$ patients in the dataset who are Pima Indian females of age $21$ and above. The record of each of them includes $8$ pieces of information: the number of times that they are pregnant, the plasma glucose concentration over $2$~hours in an oral glucose tolerance test, the blood pressure, the triceps skinfold thickness, the amount of $2$-hour serum insulin, the body mass index, the diabetes pedigree function (based on the family history), and the age. These $8$ features are used to predict whether a woman is diabetic. We use a logistic regression model for this task which has $9$ parameters including the weights for the $8$ features and a bias.

\begin{figure}[th]
    \centering
    \begin{tabular}{@{}c@{}}
        \includegraphics[width=0.39\textwidth]{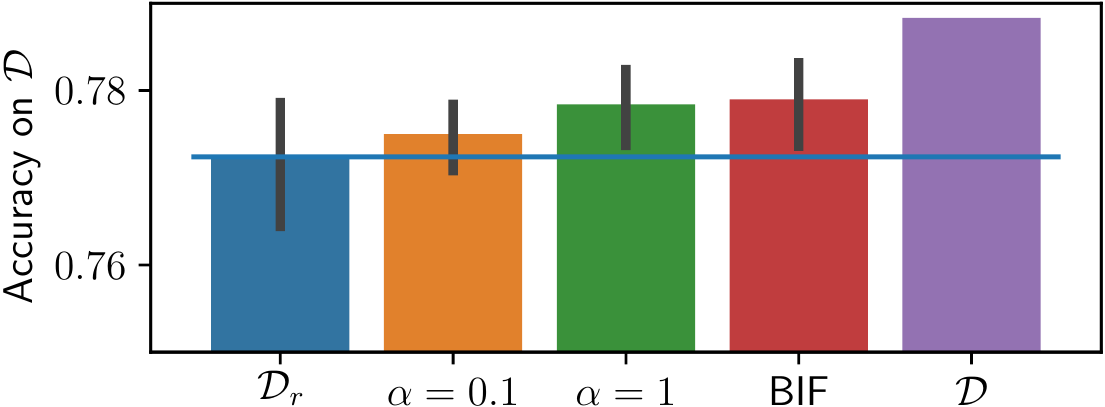}
        \\
        (a) 
        \\\vspace{-2mm}
        \\
        \includegraphics[width=0.39\textwidth]{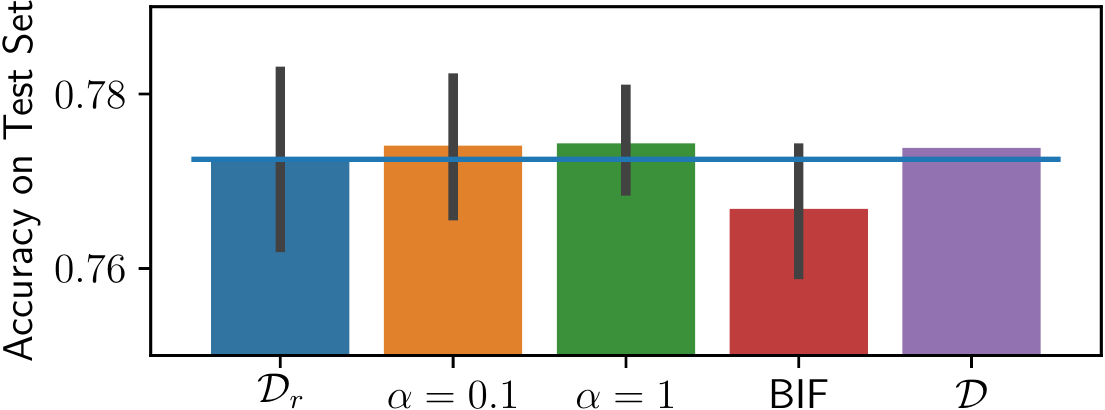}
        \\
        (b)
        \\\vspace{-2mm}
        \\
        \includegraphics[width=0.39\textwidth]{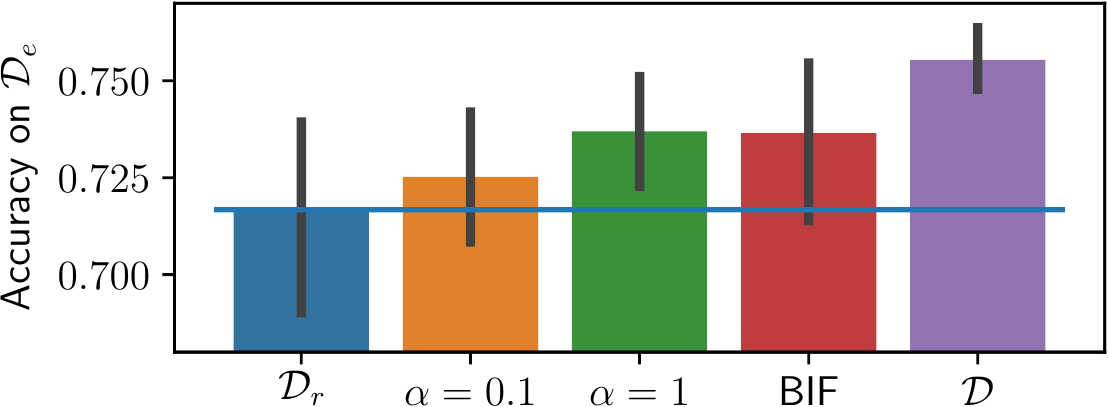}
        \\
        (c)
        \vspace{-3mm}
    \end{tabular}

    \caption{Accuracy of the model trained on $\mcl{D}$ (labeled as $\mcl{D}$), retrained on $\mcl{D}_r$ (labeled as $\mcl{D}_r$), unlearned using BIF, \acron{} with $\alpha = 1$ (labeled as $\alpha = 1$) and $\alpha = 0.1$ (labeled as $\alpha=0.1$). The accuracy is evaluated on (a) the training dataset $\mcl{D}$, (b) the test set, and (c) the erased dataset $\mcl{D}_e$. The bar plots show the average accuracy over $23$ unlearning tasks with $23$ different erased datasets $\mcl{D}_e$. The black solid lines show the $95\%$ confidence interval of the accuracy. The blue line shows the desirable accuracy (i.e., the accuracy of the retrained model) of the unlearned model.}
    \label{fig:diabetes}
    \vspace{-0.2cm}
\end{figure}

We look at a scenario where $200$ patients (i.e., $|\mcl{D}_e| = 200$) decide to withdraw their information from the trained model.
Suppose that the medical data are confidential, so after training the model on the training dataset $\mcl{D}$, only the trained model is kept (i.e., $\bm{\theta}_{\mcl{D}}$) and the training data is not accessible anymore (except for the erased data $\mcl{D}_e$ that are provided by the patients withdrawing their information).
We use the \acron{} approach which utilizes $\mcl{D}_e$ and the candidate set to remove $\mcl{D}_e$ from the trained model.

For evaluation purposes, we reserve $178$ data points from $\mcl{D}$ to construct a test set that is not used in training the model.
We select $23$ different erased datasets $\mcl{D}_e$ (each of size $200$) such that removing each of them can cause a change in the model. The aim is to ease the visualization of the performance difference between unlearning methods which include \acron{} and BIF.
The average and the $95\%$ confidence interval of the accuracy of the unlearned models, the model trained on $\mcl{D}$, and the model retrained on $\mcl{D}_r$ over these $23$ unlearning scenarios are reported. 

It is important to note in this scenario, higher accuracy is not better. Instead, we want the model to `forget' the erased data $\mcl{D}_e$. Hence, 
we would like the accuracy of the unlearned model to match that of the model retrained on $\mcl{D}_r$.

The prior distribution of each model parameter is a Gaussian distribution with mean $0$ and variance $3$. We construct the candidate set and the enlarged candidate set (of the $9$ model parameters) using $10,000$ MCMC samples.

\begin{figure}
    \centering
    \includegraphics[width=0.42\textwidth]{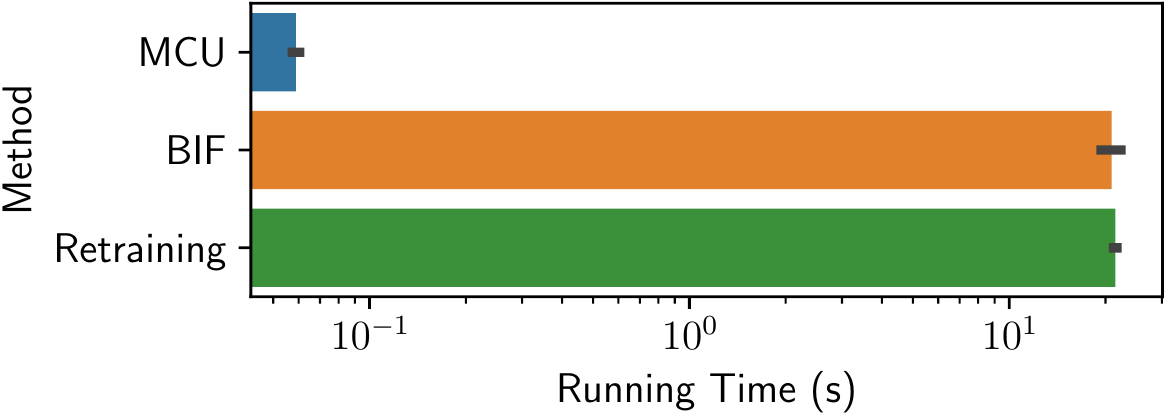}
    \vspace{-2mm}
    \caption{Running time (in seconds) for \acron{}, BIF, and retraining methods.}
    \label{fig:timediabetes}
    \vspace{-0.4cm}
\end{figure}

To empirically show the unlearning performance of \acron{} with $\alpha=1$ (not flattening), $\alpha=0.1$ (flattening), and BIF, Figure~\ref{fig:diabetes} plots the accuracy of the unlearned models using these methods on the training dataset $\mcl{D}$, the test set, and the erased dataset $\mcl{D}_e$.
As explained above, an unlearning method should be able to achieve an accuracy similar to that of the model retrained on $\mcl{D}_r$ (labeled as $\mcl{D}_r$ in Figure~\ref{fig:diabetes}). 
Intuitively, by unlearning the model trained on $\mcl{D}$ from the erased dataset $\mcl{D}_e$, the accuracy of the unlearned model on the training dataset $\mcl{D}$ and the erased dataset $\mcl{D}_e$ should decrease. It can be interpreted as the unlearned model `forgetting' the erased dataset $\mcl{D}_e$. This trend can be clearly observed in the case of the retrained model (by comparing the bar plot labeled as $\mcl{D}_r$ with the bar plot labeled as $\mcl{D}$) in Figures~\ref{fig:diabetes}a and~\ref{fig:diabetes}c.
Between our proposed \acron{} methods and BIF, our \acron{} methods, especially the one with an enlarged candidate set using a flattened distribution with $\alpha=0.1$, outperform BIF --- their accuracies on the training dataset $\mcl{D}$ and the erased dataset $\mcl{D}_e$ are closer to that of the model retrained on the remaining dataset $\mcl{D}_r$. As for the accuracy of the test set in Figure~\ref{fig:diabetes}b, while the unlearned model obtained with \acron{} is able to maintain similar accuracy to that of the retrained model, the accuracy of the unlearned model obtained with BIF drops significantly compared with the desired accuracy of the retrained model. In Figure~\ref{fig:timediabetes}, we also observe that \acron{} incurs much less time than BIF and retraining methods as it does not use the remaining dataset during the unlearning procedure.

\subsection{Phishing Webpage Detection Dataset}
\label{sec:phish}

We consider the phishing webpage dataset~\cite{phishing-dataset-MADWeb} which is used in~\cite{lee2020building} for phishing detection with supervised learning. While the phishing URLs were obtained by crawling PhishTank feed~\cite{phishtank}, the benign pages were crawled randomly from the top 300,000 websites as ranked by Alexa~\cite{alexa}.
We construct a `clean training set', denoted as $\mcl{D}^{(c)} = \{\mbf{x}_i,y_i\}_{i=1}^{100000}$, of size $100,000$ and a test set of size $30,127$, where $y_i = 1$ if $\mbf{x}_i$ corresponds to (the feature vector of) a phishing webpage and $y_i=0$ otherwise. The total number of features in the processed dataset is $52$. We also carry out experiments with a smaller subset of $5$ features that are selected based on its importance score obtained from the trained model. A logistic regression model is employed to classify the phishing webpages.

\subsubsection{Unlearning Erroneous Data}

We consider a scenario where the training dataset $\mcl{D}$ contains a subset of erroneous data (i.e., the erased data $\mcl{D}_e \subset \mcl{D}$) that has a negative impact on the model performance. 
We called this training dataset the `corrupt training dataset' to distinguish from the clean training dataset ($\mcl{D}^{(c)}$) that contains all correct labels.
The goal is to unlearn the model trained on $\mcl{D}$ from the erased data $\mcl{D}_e$.

To evaluate the performance of \acron{}, we perform experiments on $20$ different corrupt datasets and their corresponding erased datasets. Each corrupt dataset $\mcl{D}$ is constructed from the clean training dataset $\mcl{D}^{(c)}$ as follows:
\begin{enumerate}[leftmargin=*]
\item A data point (i.e., a vector of feature values extracted from a webpage) $\mbf{x}_i$ is randomly selected from the clean training dataset $\mcl{D}^{(c)}$. Then, a set $\mathcal{N}(\mbf{x}_i)$ is formed, which consists of $200$ nearest neighbors of $\mbf{x}_i$ in~$\mcl{D}_c$ (including $\mbf{x}_i$). As the erased dataset $\mcl{D}_e$ is assumed to be erroneous, $\mathcal{N}(\mbf{x}_i)$ is chosen to form $\mcl{D}_e$, such that each point's label in $\mcl{D}_e$ is $1 - y$ if it is labeled as $y$ in~$\mcl{D}^{(c)}$. In other words, all data points in $\mcl{D}_e$ are incorrectly labeled.

\item The remaining dataset $\mcl{D}_r$ is defined as the largest subset of $\mcl{D}^{(c)}$ that does not have any common data points with $\mathcal{N}(\mbf{x}_i)$. It implies $\mcl{D}_r \cap \mcl{D}_e = \emptyset$.

\item The corrupt training dataset is defined as $\mcl{D} \triangleq \mcl{D}_r \cup \mcl{D}_e$ which consists of the clean data points from $\mcl{D}_r$ and the erroneous data points from $\mcl{D}_e$.
\end{enumerate}

\begin{figure}[t]
    \centering
    \begin{tabular}{@{}cc@{}}
    \includegraphics[width=0.2\textwidth]{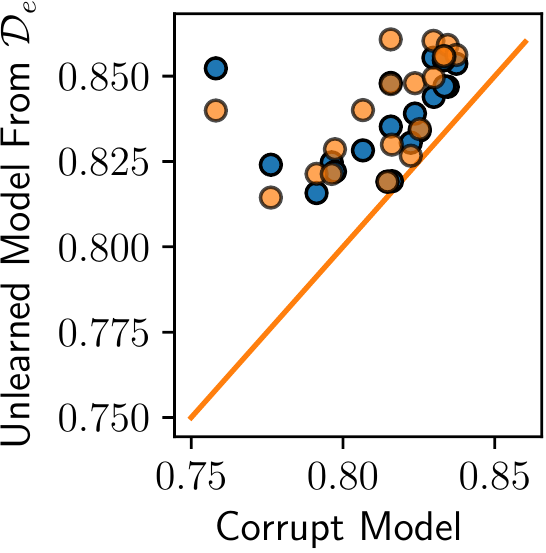}
    &
    \includegraphics[width=0.2\textwidth]{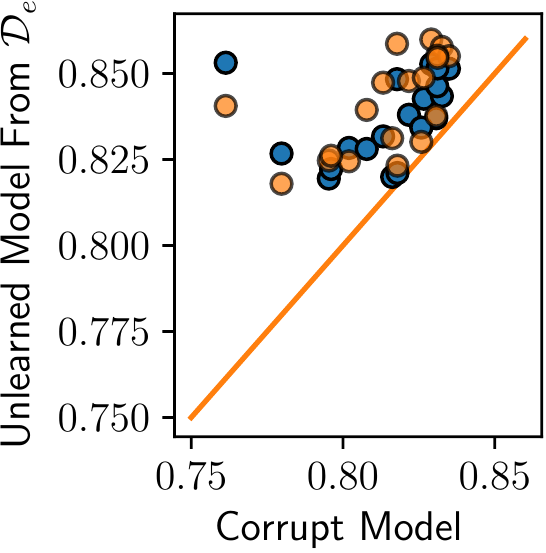}
    \\
    (a) Accuracy of $\mcl{D}^{(c)}$.
    &
    (b) Accuracy of test set.
    \\
    \\
    \includegraphics[width=0.2\textwidth]{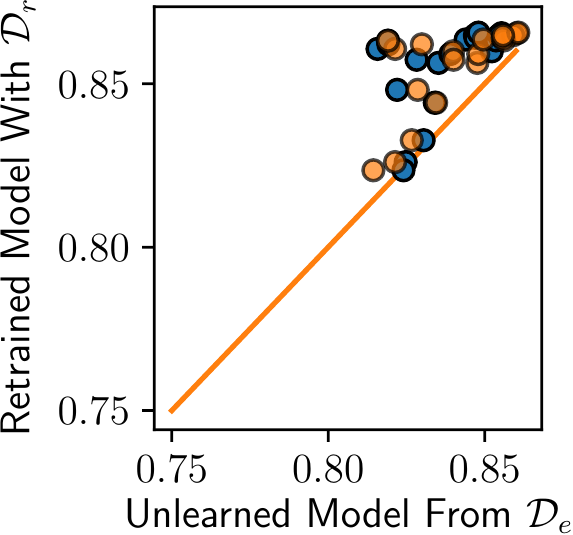}
    &
    \includegraphics[width=0.2\textwidth]{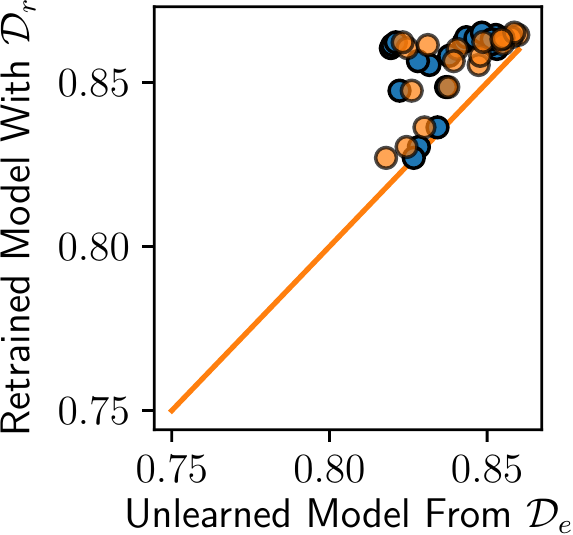}
    \\
    (c) Accuracy of $\mcl{D}^{(c)}$.
    &
    (d) Accuracy of test set.
    \end{tabular}
    \caption{Performances of \acron{} approach with $\alpha=0.01$ (shown as blue dots) and BIF (shown as orange dots), in the experiments on the phishing webpage detection dataset with $5$ features. The orange line is where the accuracy in the horizontal axis is equal to that in the vertical axis.}
    \label{fig:phish5}
    \vspace{-0.2cm}
\end{figure}

Each dot in Figure~\ref{fig:phish5} shows the result of the experiment with one corrupt dataset. The result of BIF is shown with orange dots and that of our proposed approach \acron{} with $\alpha=0.01$ is shown with blue dots. We have the following observations for the experiments on the phishing webpage detection dataset with $5$ features:
\begin{itemize}[leftmargin=*]
\item Accuracy of the unlearned model vs. that of the corrupt model: Figure~\ref{fig:phish5}a plots the accuracy of $\mcl{D}^{(c)}$ of the corrupt model (i.e., the model trained on $\mcl{D}$) against the accuracy of $\mcl{D}^{(c)}$ of the model unlearned from $\mcl{D}_e$. As $\mcl{D}_e$ is the erroneous part of the $\mcl{D}$, we expect that the accuracy of the unlearned model should improve after unlearning. This is observed in Figure~\ref{fig:phish5}a for both \acron{} with $\alpha=0.01$ and BIF (the orange and blue dots are above the orange line where the accuracy of the corrupt model is equal to that of the unlearned model). Similarly, the same observation can be made from Figure~\ref{fig:phish5}b for the accuracy of the test set.
\item Accuracy of the unlearned model vs. that of the retrained model: Figure~\ref{fig:phish5}c plots the accuracy of $\mcl{D}^{(c)}$ of the model unlearned from $\mcl{D}_e$ against the accuracy of $\mcl{D}^{(c)}$ of the model retrained on $\mcl{D}_r$. The better the unlearning approach is, the closer the accuracy of the unlearned model is to that of the naively retrained model. In Figure~\ref{fig:phish5}c, we observe that \acron{} and BIF have similar performance (note the distance between the blue and orange dots to the orange line where the accuracy of the unlearned model is equal to that of the retrained model). We have a similar observation from Figure~\ref{fig:phish5}d for the accuracy of the test set.
\end{itemize}

\begin{figure}[t]
    \centering
    \begin{tabular}{@{}cc@{}}
    \includegraphics[width=0.2\textwidth]{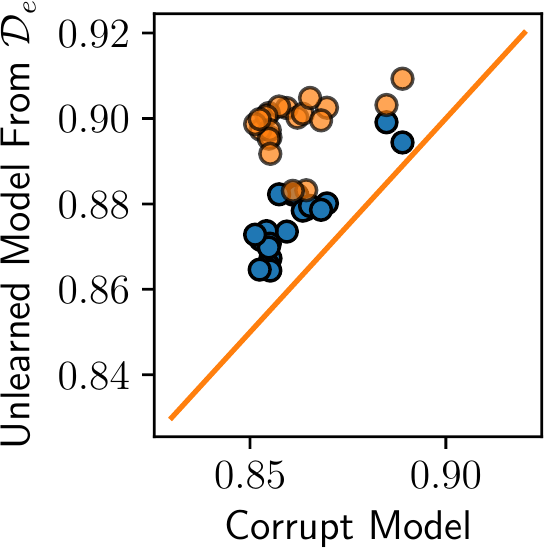}
    &
    \includegraphics[width=0.2\textwidth]{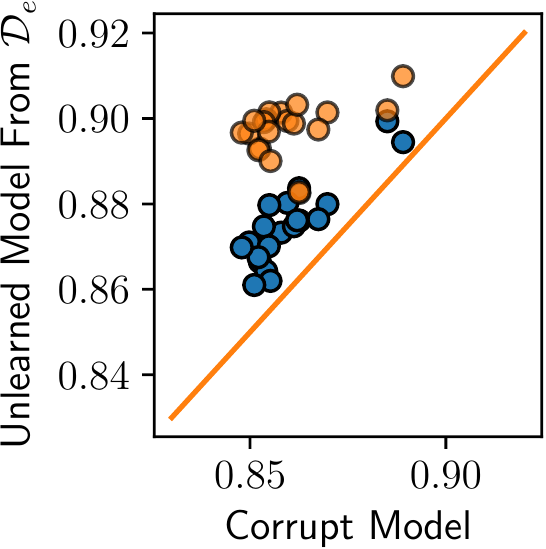}
    \\
    (a) Accuracy of $\mcl{D}^{(c)}$.
    &
    (b) Accuracy of test set.
    \\
    \\
    \includegraphics[width=0.2\textwidth]{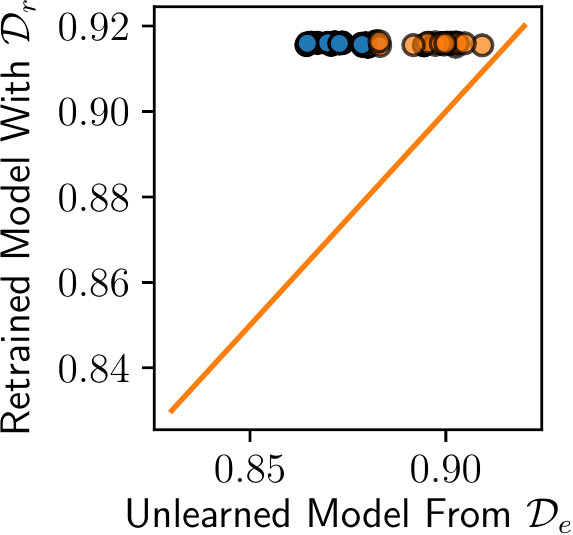}
    &
    \includegraphics[width=0.2\textwidth]{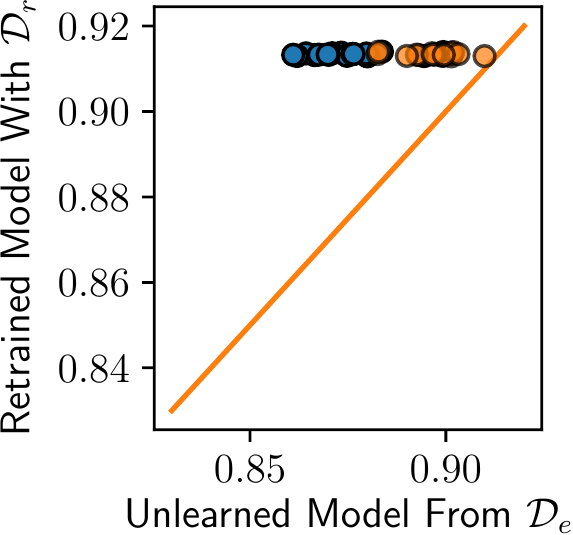}
    \\
    (c) Accuracy of $\mcl{D}^{(c)}$.
    &
    (d) Accuracy of test set.
    \end{tabular}
    \caption{Plots of the performances of our \acron{} approach with $\alpha=0.01$ (shown as blue dots) and BIF (shown as orange dots) in the experiments on the phishing webpage detection dataset with $52$ features. The orange line is where the accuracy in the horizontal axis is equal to that in the vertical axis.}
    \label{fig:phish52}
    \vspace{-0.3cm}
\end{figure}

Figure~\ref{fig:phish52} shows similar performance (comparing the accuracy of the corrupt and the retrained models on $\mcl{D}^{(c)}$ and the test set) for the experiments on the phishing webpage detection dataset with $52$ features. While BIF performance is better than \acron{} in this figure, recall that BIF uses the remaining dataset $\mcl{D}_r$ in the unlearning procedure while \acron{} does not. Furthermore, our \acron{} approach demonstrates a reasonable performance as the accuracy of the unlearned model improves over that of the corrupt model (in Figures~\ref{fig:phish52}a and~\ref{fig:phish52}b).

\subsubsection{Explaining the Effect of a Subset of Training Data on the Model Prediction} Consider another scenario where we would like to examine several subsets of training data to identify a subset that may have a negative impact on the model performance. These subsets may come from different sources such as different companies or different crowdworkers (see Section~\ref{sec:datainfluence}).
We make use of our unlearning approach \acron{} to measure the effect of a subset of training data on the model prediction, by comparing the accuracy of the model before and after the unlearning procedure. We expect that the accuracy of the model after unlearning from a subset of erroneous data should improve, while the accuracy of the model after unlearning from a subset of clean data should drop.

To construct the ground truth dataset, we make use of the above corrupt training dataset $\mcl{D}$. For a corrupt training dataset $\mcl{D}$, the corresponding erased dataset $\mcl{D}_e$ is the subset of the data that contains erroneous labels, i.e., $\mcl{D}_e$ has a negative impact on the model performance. Erasing $\mcl{D}_e$ should improve the model performance. We further draw randomly $30$ subsets of the same size as $\mcl{D}_e$ ($200$ data points) from $\mcl{D}_r$. These $30$ subsets contain correct labels. Erasing such subsets of data should make the model performance drop.

\begin{figure}[t]
    \centering
    \begin{tabular}{@{}cc@{}}
    \multicolumn{2}{c}{\includegraphics[width=0.3\textwidth]{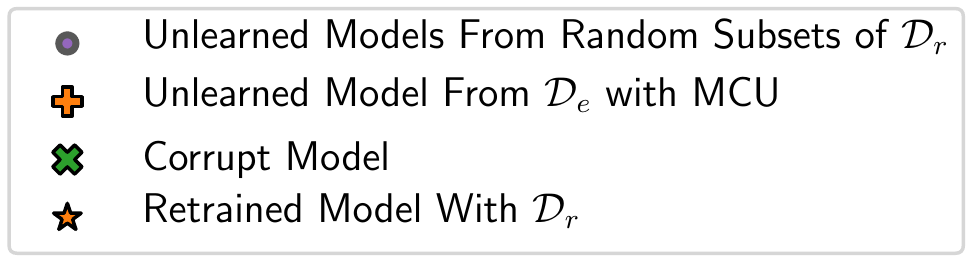}}
    \\
    \includegraphics[height=0.105\textwidth]{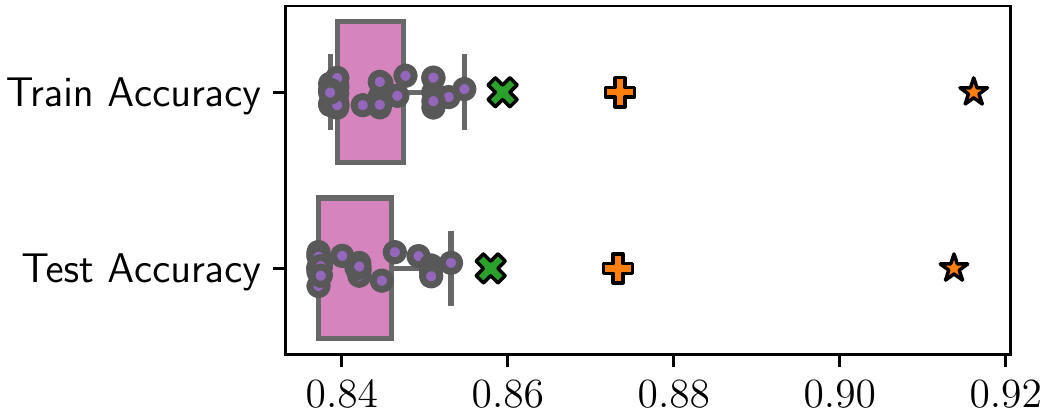}
    &
    \includegraphics[height=0.105\textwidth]{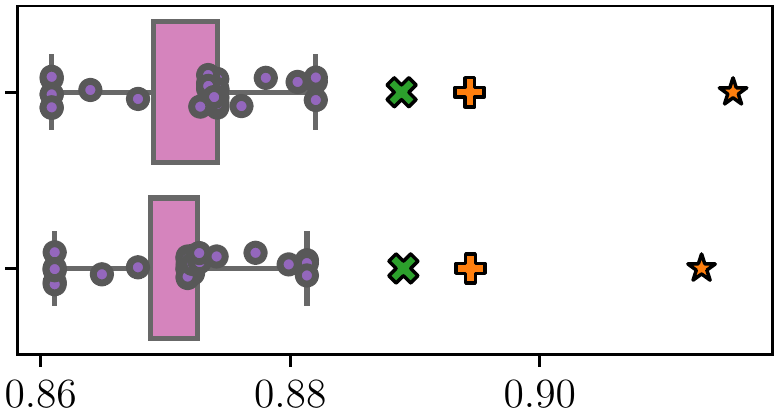}
    \\ 
    \includegraphics[height=0.105\textwidth]{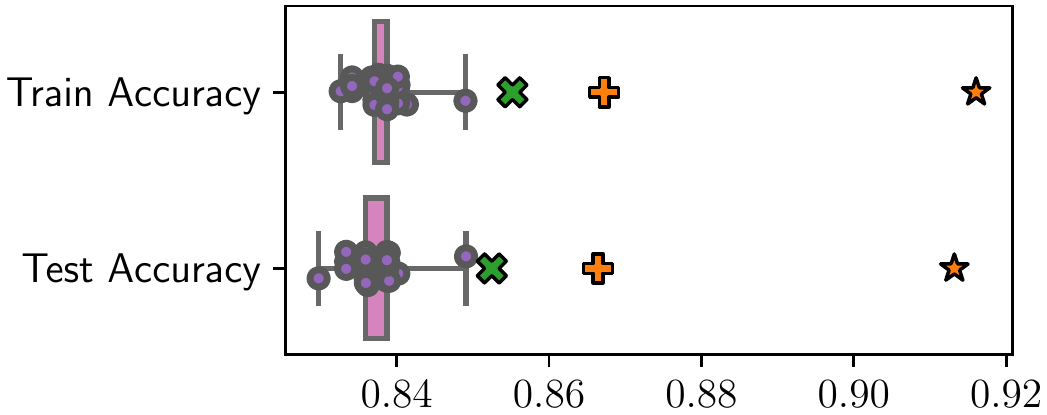}
    &
    \includegraphics[height=0.105\textwidth]{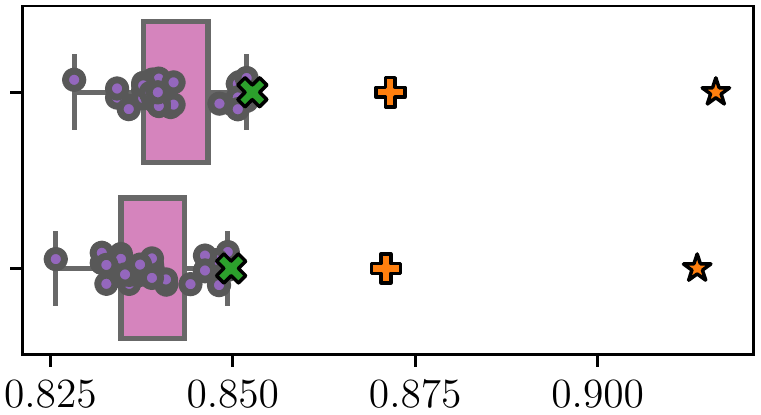}
    \end{tabular}
    \caption{Plots of the accuracies of the models unlearned from 4 different $\mcl{D}_e$ and random subsets of $\mcl{D}_r$. They are computed on the clean training set $\mcl{D}^{(c)}$ (labeled as train accuracy) and the test set (labeled as test accuracy) of the phishing webpage detection dataset with $52$ features.}
    \label{fig:phish52influence}
    \vspace{-0.2cm}
\end{figure}

In Figure~\ref{fig:phish52influence}, there are $4$ plots corresponding to $4$ corrupt training datasets that are randomly generated. Given a corrupt training dataset, the plot shows the accuracies of models unlearned from $\mcl{D}_e$ (shown as plus markers) and from random subsets of $\mcl{D}_r$ (shown as purple dots).
The results show that $\mcl{D}_e$ has a negative impact on the model prediction and its removal improves accuracy (from cross to plus).
On the other hand, removing correctly label data results in a drop in accuracy (from cross to dots).
Furthermore, by observing the boxplot of the accuracies of models unlearned from random subsets of $\mcl{D}_r$ (shown as a purple box-and-whisker diagram in the plots), we observe that the accuracies of both the corrupt model and the model unlearned from $\mcl{D}_e$ are significantly higher than that of the model unlearned from random subsets of $\mcl{D}_r$ (the plus and cross markers are outside the boxplot). 

In summary, \acron{} is able to identify subset of data which has a negative or positive impact on the original dataset by observing how its removal impacts accuracy. This capability of \acron{} is useful in detecting, identifying and (subsequently) removing adversarial data injected by an attacker.

\setlength{\tabcolsep}{1pt}

\section{Conclusion}

We propose a new machine unlearning approach, named \acron{}, to remove the effect of a specific subset of training data on the trained model. It has important applications in ensuring the ``right to be forgotten'' in the context of user privacy, erasing a subset of malicious or adversarial data from the model, and explaining the effect of a subset of training data on the model. The approach shows promising empirical performance on both synthetic datasets and real-world datasets (the Pima Indians diabetes dataset and a phishing webpage detection dataset). As next steps, we plan to consider combining this approach with the influence function, developing an online unlearning variant~\cite{yizhou22}, and also investigating these problems for unsupervised learning models.

\section*{Acknowledgement}

This research is supported by the National Research Foundation, Singapore under its Strategic Capability Research Centres Funding Initiative, and the National Research Foundation, Prime Minister’s Office, Singapore under its Corporate Laboratory@University Scheme, National University of Singapore, and Singapore Telecommunications Ltd. Any opinions, findings, and conclusions or recommendations expressed in this material are those of the author(s) and do not reflect the views of National Research Foundation, Singapore.

\vspace{-2mm}
\bibliographystyle{ACM-Reference-Format}
\bibliography{main}


\begin{thebibliography}{60}


\ifx \showCODEN    \undefined \def \showCODEN     #1{\unskip}     \fi
\ifx \showDOI      \undefined \def \showDOI       #1{#1}\fi
\ifx \showISBNx    \undefined \def \showISBNx     #1{\unskip}     \fi
\ifx \showISBNxiii \undefined \def \showISBNxiii  #1{\unskip}     \fi
\ifx \showISSN     \undefined \def \showISSN      #1{\unskip}     \fi
\ifx \showLCCN     \undefined \def \showLCCN      #1{\unskip}     \fi
\ifx \shownote     \undefined \def \shownote      #1{#1}          \fi
\ifx \showarticletitle \undefined \def \showarticletitle #1{#1}   \fi
\ifx \showURL      \undefined \def \showURL       {\relax}        \fi
\providecommand\bibfield[2]{#2}
\providecommand\bibinfo[2]{#2}
\providecommand\natexlab[1]{#1}
\providecommand\showeprint[2][]{arXiv:#2}

\bibitem[\protect\citeauthoryear{??}{fir}{2018}]%
        {fireEye-ML-mw-2018}
 \bibinfo{year}{2018}\natexlab{}.
\newblock \bibinfo{booktitle}{\emph{{MalwareGuard: FireEye’s Machine Learning
  Model to Detect and Prevent Malware}}}.
\newblock
\urldef\tempurl%
\url{https://www.fireeye.com/blog/products-and-services/2018/07/malwareguard-fireeye-machine-learning-model-to-detect-and-prevent-malware.html}
\showURL{%
\tempurl}


\bibitem[\protect\citeauthoryear{??}{phi}{2019}]%
        {phishing-dataset-MADWeb}
 \bibinfo{year}{2019}\natexlab{}.
\newblock \bibinfo{booktitle}{\emph{{Phishing webpage detection dataset}}}.
\newblock
\urldef\tempurl%
\url{https://github.com/JehLeeKR/phishing-madweb/}
\showURL{%
\tempurl}


\bibitem[\protect\citeauthoryear{??}{ale}{2021}]%
        {alexa}
 \bibinfo{year}{[Accessed: Nov. 2021]}\natexlab{}.
\newblock \bibinfo{booktitle}{\emph{{Alexa top rank websites}}}.
\newblock
\urldef\tempurl%
\url{https://www.alexa.com/topsites}
\showURL{%
\tempurl}


\bibitem[\protect\citeauthoryear{??}{ano}{2021}]%
        {anomali}
 \bibinfo{year}{[Accessed: Nov. 2021]}\natexlab{}.
\newblock \bibinfo{booktitle}{\emph{{Anomali}}}.
\newblock
\urldef\tempurl%
\url{https://www.anomali.com/marketplace/threat-intelligence-feeds}
\showURL{%
\tempurl}


\bibitem[\protect\citeauthoryear{??}{APW}{2021}]%
        {APWG}
 \bibinfo{year}{[Accessed: Nov. 2021]}\natexlab{}.
\newblock \bibinfo{booktitle}{\emph{{APWG}}}.
\newblock
\urldef\tempurl%
\url{https://apwg.org/}
\showURL{%
\tempurl}


\bibitem[\protect\citeauthoryear{??}{ope}{2021}]%
        {openphish}
 \bibinfo{year}{[Accessed: Nov. 2021]}\natexlab{}.
\newblock \bibinfo{booktitle}{\emph{{OpenPhish}}}.
\newblock
\urldef\tempurl%
\url{https://www.openphish.com/}
\showURL{%
\tempurl}


\bibitem[\protect\citeauthoryear{??}{phi}{2021}]%
        {phishtank}
 \bibinfo{year}{[Accessed: Nov. 2021]}\natexlab{}.
\newblock \bibinfo{title}{{PhishTank}}.
\newblock
\newblock
\urldef\tempurl%
\url{https://www.phishtank.com/}
\showURL{%
\tempurl}


\bibitem[\protect\citeauthoryear{??}{url}{2021}]%
        {urlhaus}
 \bibinfo{year}{[Accessed: Nov. 2021]}\natexlab{}.
\newblock \bibinfo{booktitle}{\emph{{URLhaus}}}.
\newblock
\urldef\tempurl%
\url{https://urlhaus.abuse.ch/api/}
\showURL{%
\tempurl}


\bibitem[\protect\citeauthoryear{Abdelnabi, Krombholz, and Fritz}{Abdelnabi
  et~al\mbox{.}}{2020}]%
        {visualphishnet-2020}
\bibfield{author}{\bibinfo{person}{Sahar Abdelnabi}, \bibinfo{person}{Katharina
  Krombholz}, {and} \bibinfo{person}{Mario Fritz}.}
  \bibinfo{year}{2020}\natexlab{}.
\newblock \showarticletitle{VisualPhishNet: Zero-Day Phishing Website Detection
  by Visual Similarity}. In \bibinfo{booktitle}{\emph{Proc. ACM CCS}}.
  \bibinfo{pages}{1681--1698}.
\newblock


\bibitem[\protect\citeauthoryear{Anderson and McGrew}{Anderson and
  McGrew}{2016}]%
        {enc-mw-traffic-2016}
\bibfield{author}{\bibinfo{person}{Blake Anderson} {and} \bibinfo{person}{David
  McGrew}.} \bibinfo{year}{2016}\natexlab{}.
\newblock \showarticletitle{Identifying Encrypted Malware Traffic with
  Contextual Flow Data}. In \bibinfo{booktitle}{\emph{Proc. ACM Workshop on
  Artificial Intelligence and Security}} \emph{(\bibinfo{series}{AISec '16})}.
  \bibinfo{pages}{35–46}.
\newblock


\bibitem[\protect\citeauthoryear{Anderson, Quist, Neil, Storlie, and
  Lane}{Anderson et~al\mbox{.}}{2011}]%
        {graph-mw-dyn-2011}
\bibfield{author}{\bibinfo{person}{Blake Anderson}, \bibinfo{person}{Daniel
  Quist}, \bibinfo{person}{Joshua Neil}, \bibinfo{person}{Curtis Storlie},
  {and} \bibinfo{person}{Terran Lane}.} \bibinfo{year}{2011}\natexlab{}.
\newblock \showarticletitle{{Graph-based malware detection using dynamic
  analysis}}.
\newblock \bibinfo{journal}{\emph{Journal in computer Virology}}
  \bibinfo{volume}{7}, \bibinfo{number}{4} (\bibinfo{year}{2011}),
  \bibinfo{pages}{247--258}.
\newblock


\bibitem[\protect\citeauthoryear{Anderson and Roth}{Anderson and Roth}{2018}]%
        {EMBER-2018}
\bibfield{author}{\bibinfo{person}{Hyrum~S. Anderson} {and}
  \bibinfo{person}{Phil Roth}.} \bibinfo{year}{2018}\natexlab{}.
\newblock \showarticletitle{{EMBER:} An Open Dataset for Training Static {PE}
  Malware Machine Learning Models}.
\newblock \bibinfo{journal}{\emph{CoRR}}  \bibinfo{volume}{abs/1804.04637}
  (\bibinfo{year}{2018}).
\newblock
\showeprint[arxiv]{1804.04637}
\urldef\tempurl%
\url{http://arxiv.org/abs/1804.04637}
\showURL{%
\tempurl}


\bibitem[\protect\citeauthoryear{Angluin and Laird}{Angluin and Laird}{1988}]%
        {angluin1988learning}
\bibfield{author}{\bibinfo{person}{Dana Angluin} {and} \bibinfo{person}{Philip
  Laird}.} \bibinfo{year}{1988}\natexlab{}.
\newblock \showarticletitle{Learning from noisy examples}.
\newblock \bibinfo{journal}{\emph{Machine Learning}} \bibinfo{volume}{2},
  \bibinfo{number}{4} (\bibinfo{year}{1988}), \bibinfo{pages}{343--370}.
\newblock


\bibitem[\protect\citeauthoryear{Barbos, Caron, Giovannelli, and Doucet}{Barbos
  et~al\mbox{.}}{2017}]%
        {barbos17}
\bibfield{author}{\bibinfo{person}{Andrei-Cristian Barbos},
  \bibinfo{person}{Francois Caron}, \bibinfo{person}{Jean-Fran\c{c}ois
  Giovannelli}, {and} \bibinfo{person}{Arnaud Doucet}.}
  \bibinfo{year}{2017}\natexlab{}.
\newblock \showarticletitle{Clone MCMC: Parallel High-Dimensional Gaussian
  Gibbs Sampling}. In \bibinfo{booktitle}{\emph{In Proc. {NeurIPS}}},
  Vol.~\bibinfo{volume}{30}. \bibinfo{publisher}{Curran Associates, Inc.}
\newblock


\bibitem[\protect\citeauthoryear{Bell and Komisarczuk}{Bell and
  Komisarczuk}{2020}]%
        {phish-BL-analysis-2020}
\bibfield{author}{\bibinfo{person}{Simon Bell} {and} \bibinfo{person}{Peter
  Komisarczuk}.} \bibinfo{year}{2020}\natexlab{}.
\newblock \showarticletitle{{An Analysis of Phishing Blacklists: Google Safe
  Browsing, OpenPhish, and PhishTank}}. In \bibinfo{booktitle}{\emph{Proc.
  Australasian Computer Science Week Multiconference}}.
\newblock


\bibitem[\protect\citeauthoryear{Betancourt}{Betancourt}{2017}]%
        {betancourt2017conceptual}
\bibfield{author}{\bibinfo{person}{Michael Betancourt}.}
  \bibinfo{year}{2017}\natexlab{}.
\newblock \showarticletitle{A conceptual introduction to Hamiltonian Monte
  Carlo}.
\newblock \bibinfo{journal}{\emph{arXiv preprint arXiv:1701.02434}}
  (\bibinfo{year}{2017}).
\newblock


\bibitem[\protect\citeauthoryear{Bourtoule, Chandrasekaran, Choquette-Choo,
  Jia, Travers, Zhang, Lie, and Papernot}{Bourtoule et~al\mbox{.}}{2019}]%
        {bourtoule2019machine}
\bibfield{author}{\bibinfo{person}{Lucas Bourtoule}, \bibinfo{person}{Varun
  Chandrasekaran}, \bibinfo{person}{Christopher Choquette-Choo},
  \bibinfo{person}{Hengrui Jia}, \bibinfo{person}{Adelin Travers},
  \bibinfo{person}{Baiwu Zhang}, \bibinfo{person}{David Lie}, {and}
  \bibinfo{person}{Nicolas Papernot}.} \bibinfo{year}{2019}\natexlab{}.
\newblock \showarticletitle{Machine Unlearning}.
\newblock \bibinfo{journal}{\emph{arXiv preprint arXiv:1912.03817}}
  (\bibinfo{year}{2019}).
\newblock


\bibitem[\protect\citeauthoryear{Brooks, Gelman, Jones, and Meng}{Brooks
  et~al\mbox{.}}{2011}]%
        {brooks2011handbook}
\bibfield{author}{\bibinfo{person}{Steve Brooks}, \bibinfo{person}{Andrew
  Gelman}, \bibinfo{person}{Galin Jones}, {and} \bibinfo{person}{Xiao-Li
  Meng}.} \bibinfo{year}{2011}\natexlab{}.
\newblock \bibinfo{booktitle}{\emph{Handbook of Markov chain Monte Carlo}}.
\newblock \bibinfo{publisher}{CRC press}.
\newblock


\bibitem[\protect\citeauthoryear{Cao and Yang}{Cao and Yang}{2015}]%
        {cao2015towards}
\bibfield{author}{\bibinfo{person}{Yinzhi Cao} {and} \bibinfo{person}{Junfeng
  Yang}.} \bibinfo{year}{2015}\natexlab{}.
\newblock \showarticletitle{Towards making systems forget with machine
  unlearning}. In \bibinfo{booktitle}{\emph{Proc. IEEE Symposium on Security
  and Privacy}}. \bibinfo{pages}{463--480}.
\newblock


\bibitem[\protect\citeauthoryear{Chen, Fox, and Guestrin}{Chen
  et~al\mbox{.}}{2014}]%
        {chen2014stochastic}
\bibfield{author}{\bibinfo{person}{Tianqi Chen}, \bibinfo{person}{Emily Fox},
  {and} \bibinfo{person}{Carlos Guestrin}.} \bibinfo{year}{2014}\natexlab{}.
\newblock \showarticletitle{Stochastic gradient {Hamiltonian} {Monte} {Carlo}}.
  In \bibinfo{booktitle}{\emph{Proc. ICML}}. \bibinfo{pages}{1683--1691}.
\newblock


\bibitem[\protect\citeauthoryear{Chen, Zhang, and Low}{Chen
  et~al\mbox{.}}{2022}]%
        {yizhou22}
\bibfield{author}{\bibinfo{person}{Yizhou Chen}, \bibinfo{person}{Shizhuo
  Zhang}, {and} \bibinfo{person}{Bryan Kian~Hsiang Low}.}
  \bibinfo{year}{2022}\natexlab{}.
\newblock \showarticletitle{Near-Optimal Task Selection for Meta-Learning with
  Mutual Information and Online Variational Bayesian Unlearning}. In
  \bibinfo{booktitle}{\emph{Proc. AISTATS}}.
\newblock


\bibitem[\protect\citeauthoryear{Cobb and Jalaian}{Cobb and Jalaian}{2020}]%
        {cobb2020scaling}
\bibfield{author}{\bibinfo{person}{Adam~D Cobb} {and} \bibinfo{person}{Brian
  Jalaian}.} \bibinfo{year}{2020}\natexlab{}.
\newblock \showarticletitle{Scaling {Hamiltonian} {Monte} {Carlo} inference for
  {Bayesian} neural networks with symmetric splitting}.
\newblock \bibinfo{journal}{\emph{arXiv preprint arXiv:2010.06772}}
  (\bibinfo{year}{2020}).
\newblock


\bibitem[\protect\citeauthoryear{Cook and Weisberg}{Cook and Weisberg}{1982}]%
        {cook1982residuals}
\bibfield{author}{\bibinfo{person}{R~Dennis Cook} {and}
  \bibinfo{person}{Sanford Weisberg}.} \bibinfo{year}{1982}\natexlab{}.
\newblock \bibinfo{booktitle}{\emph{Residuals and influence in regression}}.
\newblock \bibinfo{publisher}{New York: Chapman and Hall}.
\newblock


\bibitem[\protect\citeauthoryear{Du, Chen, Liu, Oak, and Song}{Du
  et~al\mbox{.}}{2019}]%
        {du2019lifelong}
\bibfield{author}{\bibinfo{person}{Min Du}, \bibinfo{person}{Zhi Chen},
  \bibinfo{person}{Chang Liu}, \bibinfo{person}{Rajvardhan Oak}, {and}
  \bibinfo{person}{Dawn Song}.} \bibinfo{year}{2019}\natexlab{}.
\newblock \showarticletitle{Lifelong Anomaly Detection Through Unlearning}. In
  \bibinfo{booktitle}{\emph{Proc. ACM CCS}}. \bibinfo{pages}{1283--1297}.
\newblock


\bibitem[\protect\citeauthoryear{Dua and Graff}{Dua and Graff}{2017}]%
        {Dua:2019}
\bibfield{author}{\bibinfo{person}{Dheeru Dua} {and} \bibinfo{person}{Casey
  Graff}.} \bibinfo{year}{2017}\natexlab{}.
\newblock \bibinfo{title}{{UCI} Machine Learning Repository}.
\newblock
\newblock
\urldef\tempurl%
\url{http://archive.ics.uci.edu/ml}
\showURL{%
\tempurl}


\bibitem[\protect\citeauthoryear{Fu, He, and Tao}{Fu et~al\mbox{.}}{2021}]%
        {fu2021bayesian}
\bibfield{author}{\bibinfo{person}{Shaopeng Fu}, \bibinfo{person}{Fengxiang
  He}, {and} \bibinfo{person}{Dacheng Tao}.} \bibinfo{year}{2021}\natexlab{}.
\newblock \showarticletitle{{Bayesian} inference forgetting}.
\newblock \bibinfo{journal}{\emph{arXiv preprint arXiv:2101.06417}}
  (\bibinfo{year}{2021}).
\newblock


\bibitem[\protect\citeauthoryear{Fu, He, and Tao}{Fu et~al\mbox{.}}{2022}]%
        {fu22}
\bibfield{author}{\bibinfo{person}{Shaopeng Fu}, \bibinfo{person}{Fengxiang
  He}, {and} \bibinfo{person}{Dacheng Tao}.} \bibinfo{year}{2022}\natexlab{}.
\newblock \showarticletitle{Knowledge removal in sampling-based Bayesian
  inference}. In \bibinfo{booktitle}{\emph{Proc. {ICLR}}}.
\newblock
\newblock
\shownote{(Part of Shaopeng Fu, Fengxiang He, and Dacheng Tao. ``{Bayesian}
  inference forgetting.'').}


\bibitem[\protect\citeauthoryear{Garg, Goldwasser, and Vasudevan}{Garg
  et~al\mbox{.}}{2020}]%
        {garg2020formalizing}
\bibfield{author}{\bibinfo{person}{Sanjam Garg}, \bibinfo{person}{Shafi
  Goldwasser}, {and} \bibinfo{person}{Prashant~Nalini Vasudevan}.}
  \bibinfo{year}{2020}\natexlab{}.
\newblock \showarticletitle{Formalizing Data Deletion in the Context of the
  Right to be Forgotten}.
\newblock \bibinfo{journal}{\emph{arXiv preprint arXiv:2002.10635}}
  (\bibinfo{year}{2020}).
\newblock


\bibitem[\protect\citeauthoryear{GDPR.EU}{GDPR.EU}{2016}]%
        {GDPR-right-to-be-forgotten}
\bibfield{author}{\bibinfo{person}{GDPR.EU}.} \bibinfo{year}{2016}\natexlab{}.
\newblock \bibinfo{title}{{Everything you need to know about the “Right to be
  forgotten”}}.
\newblock
\newblock
\urldef\tempurl%
\url{https://gdpr.eu/right-to-be-forgotten}
\showURL{%
\tempurl}


\bibitem[\protect\citeauthoryear{Ginart, Guan, Valiant, and Zou}{Ginart
  et~al\mbox{.}}{2019}]%
        {ginart2019making}
\bibfield{author}{\bibinfo{person}{Antonio Ginart}, \bibinfo{person}{Melody
  Guan}, \bibinfo{person}{Gregory Valiant}, {and} \bibinfo{person}{James~Y
  Zou}.} \bibinfo{year}{2019}\natexlab{}.
\newblock \showarticletitle{Making AI forget you: Data deletion in machine
  learning}. In \bibinfo{booktitle}{\emph{Advances in Neural Information
  Processing Systems}}. \bibinfo{pages}{3513--3526}.
\newblock


\bibitem[\protect\citeauthoryear{Goldberger and Ben-Reuven}{Goldberger and
  Ben-Reuven}{2017}]%
        {goldberger2017}
\bibfield{author}{\bibinfo{person}{Jacob Goldberger} {and}
  \bibinfo{person}{Ehud Ben-Reuven}.} \bibinfo{year}{2017}\natexlab{}.
\newblock \showarticletitle{Training deep neural-networks using a noise
  adaptation layer}. In \bibinfo{booktitle}{\emph{Proc. ICLR}}.
\newblock


\bibitem[\protect\citeauthoryear{Goodfellow, Shlens, and Szegedy}{Goodfellow
  et~al\mbox{.}}{2014}]%
        {goodfellow2014explaining}
\bibfield{author}{\bibinfo{person}{Ian~J Goodfellow}, \bibinfo{person}{Jonathon
  Shlens}, {and} \bibinfo{person}{Christian Szegedy}.}
  \bibinfo{year}{2014}\natexlab{}.
\newblock \showarticletitle{Explaining and harnessing adversarial examples}.
\newblock \bibinfo{journal}{\emph{arXiv preprint arXiv:1412.6572}}
  (\bibinfo{year}{2014}).
\newblock


\bibitem[\protect\citeauthoryear{Guo, Goldstein, Hannun, and van~der
  Maaten}{Guo et~al\mbox{.}}{2019}]%
        {guo2019certified}
\bibfield{author}{\bibinfo{person}{Chuan Guo}, \bibinfo{person}{Tom Goldstein},
  \bibinfo{person}{Awni Hannun}, {and} \bibinfo{person}{Laurens van~der
  Maaten}.} \bibinfo{year}{2019}\natexlab{}.
\newblock \showarticletitle{Certified Data Removal from Machine Learning
  Models}.
\newblock \bibinfo{journal}{\emph{arXiv preprint arXiv:1911.03030}}
  (\bibinfo{year}{2019}).
\newblock


\bibitem[\protect\citeauthoryear{Hastings}{Hastings}{1970}]%
        {hastings1970monte}
\bibfield{author}{\bibinfo{person}{W~Keith Hastings}.}
  \bibinfo{year}{1970}\natexlab{}.
\newblock \showarticletitle{{Monte} {Carlo} sampling methods using {Markov}
  chains and their applications}.
\newblock \bibinfo{journal}{\emph{Biometrika}} \bibinfo{volume}{57},
  \bibinfo{number}{1} (\bibinfo{year}{1970}), \bibinfo{pages}{97--109}.
\newblock


\bibitem[\protect\citeauthoryear{Hoffman, Gelman, et~al\mbox{.}}{Hoffman
  et~al\mbox{.}}{2014}]%
        {hoffman2014no}
\bibfield{author}{\bibinfo{person}{Matthew~D Hoffman}, \bibinfo{person}{Andrew
  Gelman}, {et~al\mbox{.}}} \bibinfo{year}{2014}\natexlab{}.
\newblock \showarticletitle{The no-u-turn sampler: adaptively setting path
  lengths in {Hamiltonian} {Monte} {Carlo}}.
\newblock \bibinfo{journal}{\emph{J. Mach. Learn. Res.}} \bibinfo{volume}{15},
  \bibinfo{number}{1} (\bibinfo{year}{2014}), \bibinfo{pages}{1593--1623}.
\newblock


\bibitem[\protect\citeauthoryear{Kearns}{Kearns}{1998}]%
        {kearns1998efficient}
\bibfield{author}{\bibinfo{person}{Michael Kearns}.}
  \bibinfo{year}{1998}\natexlab{}.
\newblock \showarticletitle{Efficient noise-tolerant learning from statistical
  queries}.
\newblock \bibinfo{journal}{\emph{Journal of the {ACM} ({JACM})}}
  \bibinfo{volume}{45}, \bibinfo{number}{6} (\bibinfo{year}{1998}),
  \bibinfo{pages}{983--1006}.
\newblock


\bibitem[\protect\citeauthoryear{Koh, Ang, Teo, and Liang}{Koh
  et~al\mbox{.}}{2019}]%
        {koh2019accuracy}
\bibfield{author}{\bibinfo{person}{Pang~Wei Koh}, \bibinfo{person}{Kai-Siang
  Ang}, \bibinfo{person}{Hubert Teo}, {and} \bibinfo{person}{Percy~S Liang}.}
  \bibinfo{year}{2019}\natexlab{}.
\newblock \showarticletitle{On the accuracy of influence functions for
  measuring group effects}. In \bibinfo{booktitle}{\emph{Proc. {NeurIPS}}}.
\newblock


\bibitem[\protect\citeauthoryear{Koh and Liang}{Koh and Liang}{2017}]%
        {koh2017understanding}
\bibfield{author}{\bibinfo{person}{Pang~Wei Koh} {and} \bibinfo{person}{Percy
  Liang}.} \bibinfo{year}{2017}\natexlab{}.
\newblock \showarticletitle{Understanding black-box predictions via influence
  functions}. In \bibinfo{booktitle}{\emph{Proc. ICML}}.
  \bibinfo{pages}{1885--1894}.
\newblock


\bibitem[\protect\citeauthoryear{Le, Pham, Sahoo, and Hoi}{Le
  et~al\mbox{.}}{2018}]%
        {le2018urlnet}
\bibfield{author}{\bibinfo{person}{Hung Le}, \bibinfo{person}{Quang Pham},
  \bibinfo{person}{Doyen Sahoo}, {and} \bibinfo{person}{Steven~CH Hoi}.}
  \bibinfo{year}{2018}\natexlab{}.
\newblock \showarticletitle{{URLNet: Learning a URL representation with deep
  learning for malicious URL detection}}.
\newblock \bibinfo{journal}{\emph{arXiv preprint arXiv:1802.03162}}
  (\bibinfo{year}{2018}).
\newblock


\bibitem[\protect\citeauthoryear{Lee, Tang, Ye, Abbasi, Hay, and Divakaran}{Lee
  et~al\mbox{.}}{2021}]%
        {d-fence-2021}
\bibfield{author}{\bibinfo{person}{Jehyun Lee}, \bibinfo{person}{Farren Tang},
  \bibinfo{person}{Pingxiao Ye}, \bibinfo{person}{Fahim Abbasi},
  \bibinfo{person}{Phil Hay}, {and} \bibinfo{person}{Dinil~Mon Divakaran}.}
  \bibinfo{year}{2021}\natexlab{}.
\newblock \showarticletitle{{D-Fence: A Flexible, Efficient, and Comprehensive
  Phishing Email Detection System}}. In \bibinfo{booktitle}{\emph{IEEE European
  Symposium on Security and Privacy (IEEE EuroS\&P)}}.
  \bibinfo{pages}{578--597}.
\newblock


\bibitem[\protect\citeauthoryear{Lee, Ye, Liu, Divakaran, and Chan}{Lee
  et~al\mbox{.}}{2020}]%
        {lee2020building}
\bibfield{author}{\bibinfo{person}{Jehyun Lee}, \bibinfo{person}{Pingxiao Ye},
  \bibinfo{person}{Ruofan Liu}, \bibinfo{person}{Dinil~Mon Divakaran}, {and}
  \bibinfo{person}{Mun~Choon Chan}.} \bibinfo{year}{2020}\natexlab{}.
\newblock \showarticletitle{Building robust phishing detection system: an
  empirical analysis}. In \bibinfo{booktitle}{\emph{NDSS Workshop on
  Measurements, Attacks, and Defenses for the Web (MADWeb)}}.
\newblock


\bibitem[\protect\citeauthoryear{Lin, Liu, Divakaran, Ng, Chan, Lu, Si, Zhang,
  and Dong}{Lin et~al\mbox{.}}{2021}]%
        {phishpedia-2021}
\bibfield{author}{\bibinfo{person}{Yun Lin}, \bibinfo{person}{Ruofan Liu},
  \bibinfo{person}{Dinil~Mon Divakaran}, \bibinfo{person}{Jun~Yang Ng},
  \bibinfo{person}{Qing~Zhou Chan}, \bibinfo{person}{Yiwen Lu},
  \bibinfo{person}{Yuxuan Si}, \bibinfo{person}{Fan Zhang}, {and}
  \bibinfo{person}{Jin~Song Dong}.} \bibinfo{year}{2021}\natexlab{}.
\newblock \showarticletitle{Phishpedia: A Hybrid Deep Learning Based Approach
  to Visually Identify Phishing Webpages}. In \bibinfo{booktitle}{\emph{30th
  {USENIX} Security Symposium}}.
\newblock


\bibitem[\protect\citeauthoryear{Marchal, Saari, Singh, and Asokan}{Marchal
  et~al\mbox{.}}{2016}]%
        {marchal2016know}
\bibfield{author}{\bibinfo{person}{Samuel Marchal}, \bibinfo{person}{Kalle
  Saari}, \bibinfo{person}{Nidhi Singh}, {and} \bibinfo{person}{N Asokan}.}
  \bibinfo{year}{2016}\natexlab{}.
\newblock \showarticletitle{{Know your phish: Novel techniques for detecting
  phishing sites and their targets}}. In \bibinfo{booktitle}{\emph{Proc. IEEE
  36th International Conference on Distributed Computing Systems (ICDCS)}}.
  \bibinfo{pages}{323--333}.
\newblock


\bibitem[\protect\citeauthoryear{Metropolis, Rosenbluth, Rosenbluth, Teller,
  and Teller}{Metropolis et~al\mbox{.}}{1953}]%
        {metropolis1953equation}
\bibfield{author}{\bibinfo{person}{Nicholas Metropolis},
  \bibinfo{person}{Arianna~W Rosenbluth}, \bibinfo{person}{Marshall~N
  Rosenbluth}, \bibinfo{person}{Augusta~H Teller}, {and}
  \bibinfo{person}{Edward Teller}.} \bibinfo{year}{1953}\natexlab{}.
\newblock \showarticletitle{Equation of state calculations by fast computing
  machines}.
\newblock \bibinfo{journal}{\emph{The journal of chemical physics}}
  \bibinfo{volume}{21}, \bibinfo{number}{6} (\bibinfo{year}{1953}),
  \bibinfo{pages}{1087--1092}.
\newblock


\bibitem[\protect\citeauthoryear{Moore and Clayton}{Moore and Clayton}{2008}]%
        {phishtank-eval-2008}
\bibfield{author}{\bibinfo{person}{Tyler Moore} {and} \bibinfo{person}{Richard
  Clayton}.} \bibinfo{year}{2008}\natexlab{}.
\newblock \showarticletitle{{Evaluating the Wisdom of Crowds in Assessing
  Phishing Websites}}. In \bibinfo{booktitle}{\emph{Financial Cryptography and
  Data Security}}. \bibinfo{pages}{16--30}.
\newblock


\bibitem[\protect\citeauthoryear{Neal et~al\mbox{.}}{Neal
  et~al\mbox{.}}{2011}]%
        {neal2011}
\bibfield{author}{\bibinfo{person}{Radford~M Neal} {et~al\mbox{.}}}
  \bibinfo{year}{2011}\natexlab{}.
\newblock \showarticletitle{{MCMC} using {Hamiltonian} dynamics}.
\newblock In \bibinfo{booktitle}{\emph{Handbook of {Markov} chain {Monte}
  {Carlo}}}. \bibinfo{publisher}{CRC Press}, Chapter~5.
\newblock


\bibitem[\protect\citeauthoryear{Nevat, Divakaran, Nagarajan, Zhang, Le, Ling,
  and Thing}{Nevat et~al\mbox{.}}{2018}]%
        {NADA-2018}
\bibfield{author}{\bibinfo{person}{Ido Nevat}, \bibinfo{person}{Dinil~Mon
  Divakaran}, \bibinfo{person}{Sai~Ganesh Nagarajan}, \bibinfo{person}{Pengfei
  Zhang}, \bibinfo{person}{Su Le}, \bibinfo{person}{Ko~Li Ling}, {and}
  \bibinfo{person}{Vrizlynn Thing}.} \bibinfo{year}{2018}\natexlab{}.
\newblock \showarticletitle{{Anomaly Detection and Attribution in Networks With
  Temporally Correlated Traffic}}.
\newblock \bibinfo{journal}{\emph{IEEE/ACM Transactions on Networking}}
  \bibinfo{volume}{26}, \bibinfo{number}{1} (\bibinfo{date}{Feb.}
  \bibinfo{year}{2018}), \bibinfo{pages}{131--144}.
\newblock


\bibitem[\protect\citeauthoryear{Nguyen, Lim, Divakaran, Low, and Chan}{Nguyen
  et~al\mbox{.}}{2019}]%
        {nguyen2019gee}
\bibfield{author}{\bibinfo{person}{Quoc~Phong Nguyen}, \bibinfo{person}{Kar~Wai
  Lim}, \bibinfo{person}{Dinil~Mon Divakaran}, \bibinfo{person}{Kian~Hsiang
  Low}, {and} \bibinfo{person}{Mun~Choon Chan}.}
  \bibinfo{year}{2019}\natexlab{}.
\newblock \showarticletitle{{GEE}: {A} gradient-based explainable variational
  autoencoder for network anomaly detection}. In
  \bibinfo{booktitle}{\emph{Proc. {IEEE} Conf. on Commun. and Network Security
  ({CNS})}}. \bibinfo{pages}{91--99}.
\newblock


\bibitem[\protect\citeauthoryear{Nguyen, Low, and Jaillet}{Nguyen
  et~al\mbox{.}}{2020}]%
        {nguyen2020variational}
\bibfield{author}{\bibinfo{person}{Quoc~Phong Nguyen}, \bibinfo{person}{Bryan
  Kian~Hsiang Low}, {and} \bibinfo{person}{Patrick Jaillet}.}
  \bibinfo{year}{2020}\natexlab{}.
\newblock \showarticletitle{Variational {Bayesian} unlearning}. In
  \bibinfo{booktitle}{\emph{Proc. {NeurIPS}}}.
\newblock


\bibitem[\protect\citeauthoryear{Oest, Zhang, Wardman, Nunes, Burgis, Zand,
  Thomas, Doup{\'e}, and Ahn}{Oest et~al\mbox{.}}{2020}]%
        {phishing-life-cycle-2020}
\bibfield{author}{\bibinfo{person}{Adam Oest}, \bibinfo{person}{Penghui Zhang},
  \bibinfo{person}{Brad Wardman}, \bibinfo{person}{Eric Nunes},
  \bibinfo{person}{Jakub Burgis}, \bibinfo{person}{Ali Zand},
  \bibinfo{person}{Kurt Thomas}, \bibinfo{person}{Adam Doup{\'e}}, {and}
  \bibinfo{person}{Gail-Joon Ahn}.} \bibinfo{year}{2020}\natexlab{}.
\newblock \showarticletitle{{Sunrise to Sunset: Analyzing the End-to-end Life
  Cycle and Effectiveness of Phishing Attacks at Scale}}. In
  \bibinfo{booktitle}{\emph{29th {USENIX} Security Symposium}}.
  \bibinfo{pages}{361--377}.
\newblock


\bibitem[\protect\citeauthoryear{Onwuzurike, Mariconti, Andriotis, Cristofaro,
  Ross, and Stringhini}{Onwuzurike et~al\mbox{.}}{2019}]%
        {MaMaDroid-2019}
\bibfield{author}{\bibinfo{person}{Lucky Onwuzurike}, \bibinfo{person}{Enrico
  Mariconti}, \bibinfo{person}{Panagiotis Andriotis},
  \bibinfo{person}{Emiliano~De Cristofaro}, \bibinfo{person}{Gordon Ross},
  {and} \bibinfo{person}{Gianluca Stringhini}.}
  \bibinfo{year}{2019}\natexlab{}.
\newblock \showarticletitle{{MaMaDroid: Detecting Android Malware by Building
  Markov Chains of Behavioral Models (Extended Version)}}.
\newblock \bibinfo{journal}{\emph{ACM Trans. Priv. Secur.}}
  \bibinfo{volume}{22}, \bibinfo{number}{2}, Article \bibinfo{articleno}{14}
  (\bibinfo{date}{April} \bibinfo{year}{2019}).
\newblock


\bibitem[\protect\citeauthoryear{Patrini, Rozza, {Krishna Menon}, Nock, and
  Qu}{Patrini et~al\mbox{.}}{2017}]%
        {patrini2017making}
\bibfield{author}{\bibinfo{person}{G. Patrini}, \bibinfo{person}{A. Rozza},
  \bibinfo{person}{A. {Krishna Menon}}, \bibinfo{person}{R. Nock}, {and}
  \bibinfo{person}{L. Qu}.} \bibinfo{year}{2017}\natexlab{}.
\newblock \showarticletitle{Making deep neural networks robust to label noise:
  a loss correction approach}. In \bibinfo{booktitle}{\emph{Proc. CVPR}}.
  \bibinfo{pages}{1944--1952}.
\newblock


\bibitem[\protect\citeauthoryear{PHISHLABS}{PHISHLABS}{2021}]%
        {phishing-compromised-2021}
\bibfield{author}{\bibinfo{person}{PHISHLABS}.}
  \bibinfo{year}{2021}\natexlab{}.
\newblock \bibinfo{title}{{Most Phishing Attacks Use Compromised Domains and
  Free Hosting}}.
\newblock
\newblock
\urldef\tempurl%
\url{https://www.phishlabs.com/blog/most-phishing-attacks-use-compromised-domains-and-free-hosting/}
\showURL{%
\tempurl}


\bibitem[\protect\citeauthoryear{Robert, Elvira, Tawn, and Wu}{Robert
  et~al\mbox{.}}{2018}]%
        {robert2018accelerating}
\bibfield{author}{\bibinfo{person}{Christian~P Robert},
  \bibinfo{person}{V{\'\i}ctor Elvira}, \bibinfo{person}{Nick Tawn}, {and}
  \bibinfo{person}{Changye Wu}.} \bibinfo{year}{2018}\natexlab{}.
\newblock \showarticletitle{Accelerating {MCMC} algorithms}.
\newblock \bibinfo{journal}{\emph{Wiley interdisciplinary reviews:
  computational statistics}} \bibinfo{volume}{10}, \bibinfo{number}{5}
  (\bibinfo{year}{2018}), \bibinfo{pages}{e1435}.
\newblock


\bibitem[\protect\citeauthoryear{Saxe and Berlin}{Saxe and Berlin}{2015}]%
        {mw-detection-MALWARE-2015}
\bibfield{author}{\bibinfo{person}{Joshua Saxe} {and}
  \bibinfo{person}{Konstantin Berlin}.} \bibinfo{year}{2015}\natexlab{}.
\newblock \showarticletitle{Deep neural network based malware detection using
  two dimensional binary program features}. In \bibinfo{booktitle}{\emph{{Proc.
  10th International Conference on Malicious and Unwanted Software
  (MALWARE)}}}. \bibinfo{pages}{11--20}.
\newblock


\bibitem[\protect\citeauthoryear{Severi, Meyer, Coull, and Oprea}{Severi
  et~al\mbox{.}}{2021}]%
        {backdoor-poisoning-malware-2021}
\bibfield{author}{\bibinfo{person}{Giorgio Severi}, \bibinfo{person}{Jim
  Meyer}, \bibinfo{person}{Scott Coull}, {and} \bibinfo{person}{Alina Oprea}.}
  \bibinfo{year}{2021}\natexlab{}.
\newblock \showarticletitle{{Explanation-Guided Backdoor Poisoning Attacks
  Against Malware Classifiers}}. In \bibinfo{booktitle}{\emph{30th {USENIX}
  Security Symposium}}.
\newblock


\bibitem[\protect\citeauthoryear{Szegedy, Zaremba, Sutskever, Bruna, Erhan,
  Goodfellow, and Fergus}{Szegedy et~al\mbox{.}}{2013}]%
        {szegedy2013intriguing}
\bibfield{author}{\bibinfo{person}{Christian Szegedy},
  \bibinfo{person}{Wojciech Zaremba}, \bibinfo{person}{Ilya Sutskever},
  \bibinfo{person}{Joan Bruna}, \bibinfo{person}{Dumitru Erhan},
  \bibinfo{person}{Ian Goodfellow}, {and} \bibinfo{person}{Rob Fergus}.}
  \bibinfo{year}{2013}\natexlab{}.
\newblock \showarticletitle{Intriguing properties of neural networks}.
\newblock \bibinfo{journal}{\emph{arXiv preprint arXiv:1312.6199}}
  (\bibinfo{year}{2013}).
\newblock


\bibitem[\protect\citeauthoryear{Villaronga, Kieseberg, and Li}{Villaronga
  et~al\mbox{.}}{2018}]%
        {right-to-be-forgotten-AI-2018}
\bibfield{author}{\bibinfo{person}{Eduard~Fosch Villaronga},
  \bibinfo{person}{Peter Kieseberg}, {and} \bibinfo{person}{Tiffany Li}.}
  \bibinfo{year}{2018}\natexlab{}.
\newblock \showarticletitle{{Humans forget, machines remember: Artificial
  intelligence and the Right to Be Forgotten}}.
\newblock \bibinfo{journal}{\emph{Computer Law \& Security Review}}
  \bibinfo{volume}{34}, \bibinfo{number}{2} (\bibinfo{year}{2018}),
  \bibinfo{pages}{304--313}.
\newblock
\showISSN{0267-3649}


\bibitem[\protect\citeauthoryear{Welling and Teh}{Welling and Teh}{2011}]%
        {welling2011bayesian}
\bibfield{author}{\bibinfo{person}{Max Welling} {and} \bibinfo{person}{Yee~W
  Teh}.} \bibinfo{year}{2011}\natexlab{}.
\newblock \showarticletitle{{Bayesian} learning via stochastic gradient
  {Langevin} dynamics}. In \bibinfo{booktitle}{\emph{Proc. ICML}}.
  \bibinfo{pages}{681--688}.
\newblock


\bibitem[\protect\citeauthoryear{Zhang, Li, Zhang, Chen, and Wilson}{Zhang
  et~al\mbox{.}}{2020}]%
        {zhang2019cyclical}
\bibfield{author}{\bibinfo{person}{Ruqi Zhang}, \bibinfo{person}{Chunyuan Li},
  \bibinfo{person}{Jianyi Zhang}, \bibinfo{person}{Changyou Chen}, {and}
  \bibinfo{person}{Andrew~Gordon Wilson}.} \bibinfo{year}{2020}\natexlab{}.
\newblock \showarticletitle{Cyclical stochastic gradient {MCMC} for {Bayesian}
  deep learning}. In \bibinfo{booktitle}{\emph{Proc. ICLR}}.
\newblock


\end{thebibliography}

\appendix

\section*{Appendix}
\subsection*{Synthetic Linear Regression}
\label{sec:linearregression}

In this experiment, we would like to fit a polynomial curve $y_x = \sum_{i=0}^4 a_i x^i$ to a linear regression dataset where the output is perturbed by a Gaussian noise of mean $0$ and variance $0.01$. 
The prior distribution of each model parameter is a Gaussian distribution of mean $0$ and variance $4$. 

\begin{figure}
    \centering
    \begin{tabular}{@{}c@{}}
    \includegraphics[height=0.06\textwidth]{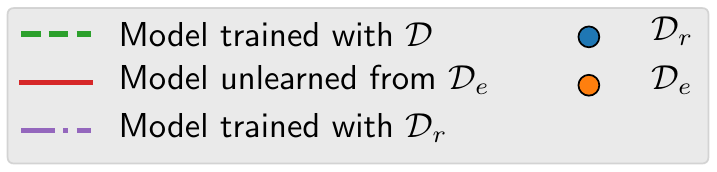}\\
    \includegraphics[width=0.3\textwidth]{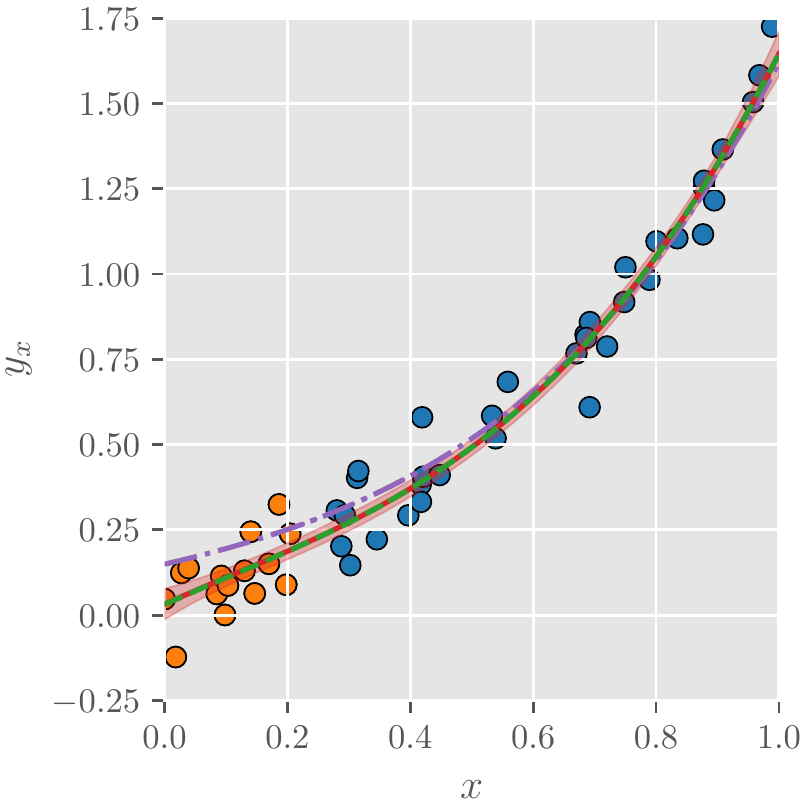}
    \\
    \makecell{(a) Model prediction from\\ unlearned model obtained by BIF.}
    \\
    \includegraphics[width=0.3\textwidth]{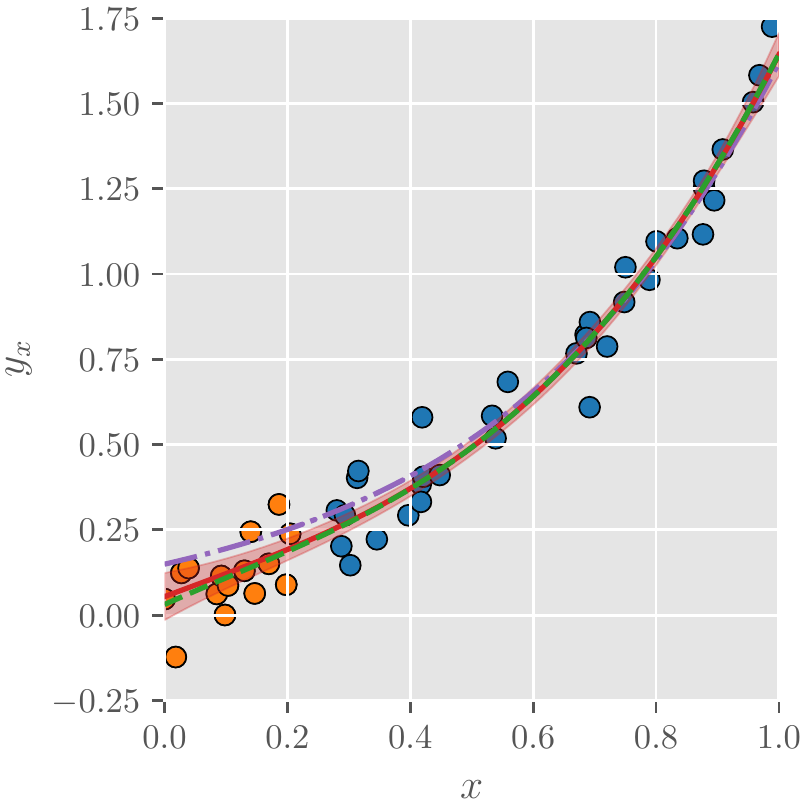}
    \\
    \makecell{(b) Model prediction from unlearned\\ model obtained by \acron{} with $\alpha = 1$.}
    \\
    \includegraphics[width=0.3\textwidth]{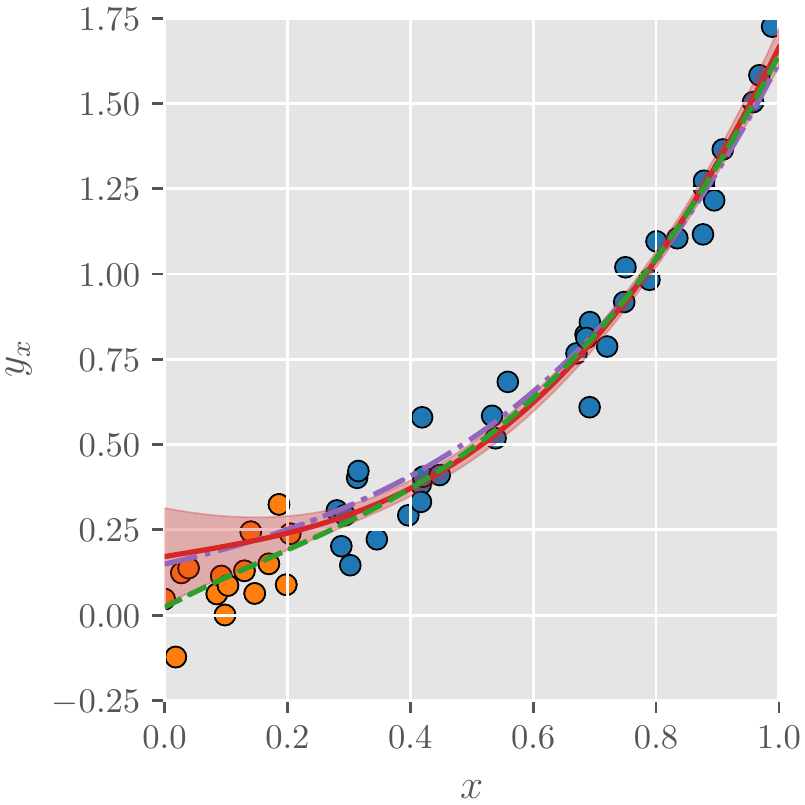}
    \\
    \makecell{(c) Model prediction from unlearned\\ model obtained by \acron{} with $\alpha=0.08$.}
    \end{tabular}
    \caption{Model prediction obtained from (a) BIF and \acron{} with (b) $\alpha=1$ and (c) $\alpha=0.08$ in experiments with the synthetic linear regression dataset.}
    \label{fig:linreg}
\end{figure}

The training dataset $\mcl{D}$ and the erased dataset $\mcl{D}_e$ consist of $50$ and $15$ data points, respectively. We observe that unlearning the model from $15$ data points randomly selected in $\mcl{D}$ often does not change the model much, i.e., the difference between $\bm{\theta}_{\mcl{D}}$ and $\bm{\theta}_{\mcl{D}_r}$ is small. Therefore, we deliberately choose $\mcl{D}_e$ to be a cluster of $15$ data points with small values of $x$ as shown in Figure~\ref{fig:linreg}.
We construct the candidate set and the enlarged candidate set (of the $5$ model parameters $\{a_i\}_{i=0}^4$) using $3000$ MCMC samples.

When the enlarged candidate set $\widetilde{\bm{\Theta}}(\alpha)$ is constructed by flattening the posterior belief $p(\bm{\theta}|\mcl{D})$ with $\tilde{p}(\bm{\theta}|\mcl{D};\alpha=0.08)$, the unlearning results are shown in Figure~\ref{fig:linreg}c. 
Compared with Figure~\ref{fig:linreg}a and~\ref{fig:linreg}b for unlearning by BIF and \acron{} with $\alpha~=~1.0$, the model prediction of the unlearned model in Figure~\ref{fig:linreg}c (plotted as a dashed green curve) is similar to that of the model retrained on $\mcl{D}_r$ (plotted as a dashed purple curve). Furthermore, we can observe that the uncertainty of the model prediction increases at the erased data (i.e., $x \in (0.0, 0.2$)) after unlearning. It can be explained by the fact that the training data in this region are erased. As a result, we observe that our proposed \acron{} method with enlarged $\widetilde{\bm{\Theta}}(\alpha)$ outperforms the other two methods in this experiment. Notably, while not using the remaining dataset, \acron{} with an enlarged candidate set outperforms BIF which uses the remaining dataset in the unlearning procedure.

\end{document}